\documentclass[10pt,a4paper]{article}

\usepackage{algorithm2e}
\usepackage{amssymb}
\usepackage{bbm}
\usepackage{dsfont}
\usepackage{amsmath}
\usepackage{subcaption}
\usepackage[font=sf,labelfont={sf,bf}, margin=1.5cm]{caption}
\usepackage[pdftex]{graphicx}
\usepackage[font=small,labelfont=bf]{caption}
\usepackage[hmargin=3cm,vmargin=3.5cm]{geometry}
\usepackage{abstract}
\usepackage{wrapfig}
\usepackage{float}
\usepackage{graphicx}
\usepackage{hyperref}
\usepackage{tikz}

\begin{document}
\title{\bf{Ensemble Committees for Stock Return Classification and Prediction}}
\author{James Brofos}
\date{\today}
\maketitle

\begin{abstract}
{\small This paper considers a portfolio trading strategy formulated
  by algorithms in the field of machine learning. The profitability of
the strategy is measured by the algorithm's capability to consistently
and accurately identify stock indices with positive or negative
returns, and to generate a preferred portfolio allocation on the basis
of a learned model. Stocks are characterized by time series data sets
consisting of technical variables that reflect market conditions in a
previous time interval, which are utilized produce binary
classification decisions in subsequent intervals. The learned model is
constructed as a committee of random forest classifiers, a non-linear
support vector machine classifier, a relevance vector machine
classifier, and a constituent ensemble of $k$-nearest neighbors
classifiers. This selection of algorithms is appealing for two
reasons: first, there is strikingly little research in economic
time-series forecasting that employs learners beyond
neural networks and clustering algorithms, and this construction
offers a viable alternative; second, this selection
incorporates an array of techniques that have both theoretically optimal
classification properties and high empirical success rates in areas
outside of finance, in
addition to offering a mixture of parametric and non-parametric
models. The ensemble committee is augmented by a boosting meta-algorithm and feature selection is performed by a
supervised Relief algorithm. The Global Industry Classification
Standard (GICS) is used to explore the ensemble model's efficacy
within the context of various fields of investment including Energy,
Materials, Financials, and Information Technology. Data from 2006 to 2012, inclusive, are considered, which are chosen for
providing a range of market circumstances for evaluating the model. The model is observed to achieve
an accuracy of approximately 70\% when predicting stock price returns
three months in advance.}
\end{abstract}

\section{Introduction}

It is crucial, in this modern era of financial
uncertainty, to explore and understand
methodologies for effectively predicting future outcomes given a
historical record. In particular, the motivating question behind this
work is whether there exists an appropriate selection of technical
explanatory variables that will produce high-accuracy stock price
return predictions. Our chosen approach to this complex financial
forecasting problem is to train an ensemble of classification
models on a subset of labeled financial data that are categorized as
members of the sets $\mathds{1}^+$ or $\mathds{1}^-$ according to
whether there was a positive or negative shift in stock price from an
initial time to a subsequent time.

This work offers two important contributions to the field of financial
forecasting. First among these is pursuant to the recommendation for
future research offered in Huerta \emph{et al.} \cite{huerta}: the learning model is
constructed such that feature selection is conducted a priori and is a
fully automated process; this allows the model itself to ``discover''
which parameters it believes are important to effective prediction
rather than being forced to accept human-designated explanatory
variables. This gives the ensemble a attribute of adaptability in the
sense that it may reformulate its relevant parameters according to the
GICS or according to location in time. Second, the construction of the
ensemble incorporates a probabilistic ranking component that
approximates a level of confidence associated with each
prediction. Within the financial literature, it is typically the case
that portfolios are formulated according to a scoring function that
is intended to capture the differences of desirable and undesirable
stocks \cite{ranking}. The capability of the ensemble is similar in
concept, yet rather than providing an absolute hierarchy of stock
preferences, a probabilistic measure of the desirability of the stock
is returned. In uncertain games such as portfolio investment, there
are clear benefits to possessing a probabilistic confidence criterion.

\section{Description of the Learning Algorithms}

This section provides a brief introduction to
the classification algorithms incorporated into the ensemble. The
section is intended for individuals with little familiarity with
learning algorithms. Notice
that this section is intended merely to introduce the constituent
models and to identify how they form a cohesive whole. Readers with
some expertise in machine learning may proceed without delay to
subsequent sections of the paper without loss of understanding. The constituent
algorithms are presented in the order (1) random forest classifier,
(2) non-linear support vector machine classifier, (3) relevance vector
machine classifier, and (4) ensemble of $k$-nearest neighbors
classifiers. The meta-algorithm of ``boosting'' is also
formalized here.

In the field of financial trading it is of some urgency to construct
models that may be learned and deployed efficiently, yet must also be
relatively robust to the inherently stochastic nature of stock
returns. Using these
crucial ideas, we can motivate an ensemble of this form by referring to
the differences between parametric and non-parametric algorithms
within machine learning. Parametric models are useful in the
sense that they are fast to learn and deploy, but typically make
strong assumptions about the distribution of the data. Non-parametric models avoid almost
all prior suppositions about the data, but the complexity of
non-parametric learning algorithms tends to be larger than that for
parameterized learners. Because random forests and $k$-nearest
neighbors classifiers are non-parametric learners, and support vector
machines and relevance vector machines are parameterized, the ensemble
model presented here benefits from the advantages of both styles, yet
still maintains the ability to deploy the constituent models
individually. 

\subsection{Random Forest}

A random forest classifier is itself a learned ensemble of decision
trees such that the constituent learners are ``decorrelated'' by
growing each tree on a randomly chosen subset of all data vectors and
all features. This ensemble results in an decision function of the
form,

\begin{align}
  f\left( \vec{x} \right)=\sum_{i=1}^m \frac{1}{m}f_i \left(
    \vec{x} \right)
\end{align}
\noindent
where $f_i$ is the decision function of the $i^{\text{th}}$ tree in
the ensemble. In growing each tree, the key idea is to minimize the
\emph{impurity} of the training data in the nodes resulting from the
candidate splits. There are several measures of such impurity
including the misclassification rate and entropy, however we elect here to
use the Gini index measurement, which is given by,

\begin{align}
  g\left( \vec{x} \right) = 1 -\sum_{c \in C} \mathbb{P} \left(
    y\vert \vec{x} = c\right)^2
\end{align}
\noindent
where $C$ is the set of potential class labels and in the binary
setting we have that $C =\{+1,-1 \}$. It can be shown that the
minimizing the Gini index at each node split is equivalent to
minimizing the expected error rate \cite{murphy}. This particular impurity
metric is chosen for the decision trees because it offers something of
a compromise between entropy and absolute error rate in terms of its
sensitivity to class probabilities. 

\subsection{Non-Linear Support Vector Machine}

The support vector machine (hereafter SVM) has remained a popular
choice in machine learning classification problems and in financial
prediction in particular \cite{huerta,kim,sewell}. This is almost certainly
because of the SVM's elegant mathematical foundations and relative
ease of implementation. The functional form of the SVM resembles,

\begin{align}
\hat{y}\left(\vec{x}\right)=\text{sign}~\left[ \hat{w}_0 +
  \sum_{i=1}^N \alpha_i y_i k\left(\vec{x},\vec{x}_i \right)\right]
\end{align}

\noindent
The constituent parameters of this formulation are described as
follows,

\begin{description}

\item[$\alpha_i$:] These are parameters that determine the shape of the
  separating hyperplane. After applying the kernel function, the data
  is ideally linearly separable in the feature space, which suggests
  why a linear construction of the hyperplane is appropriate. Learning the optimal value of the $\alpha_i$ is achieved by optimizing
the value of a quadratic program, a detailed description of which may
be found in any introductory text on machine learning. 
\item[$N$:] This is the number of support vectors that are identified
  in the optimization process used to train the SVM. The $\alpha_i$
  for non-support vector data points are zero, which is why it is only
  necessary to consider the support vectors in evaluating the
  summation.
\item[$y_i$:] These are the class labels of the support vectors, which
  all exist in the set $\{+1,-1 \}$. 
\item[$k\left(\vec{x},\vec{x}_i \right)$:] The kernel function that is
  essentially a distance augmentation. Many choices exist for the
  actual functional form of the kernel, but in this project a radial
  basis function (RBF) is used. The radial basis function resembles,
  \begin{align}
    k\left(\vec{x},\vec{x}_i \right) = e^{\gamma \vert\vert \vec{x} -
      \vec{x}_i \vert \vert ^2}
  \end{align}
\noindent
A matter that will be addressed later is the principled selection of the
parameter $\gamma$ within the RBF, which constitutes a meta-parameter
(that is, a value chosen a priori) in the model. 
\item[$\hat{w}_0$:] This is the intercept term for the hyperplane that
  separates the feature space for classification. This parameter is
  not calculated in the same way as the $\alpha_i$, but is constructed
  after the optimization process.
\end{description}

\subsection{Relevance Vector Machine}

The relevance vector machine (RVM) assumes a form similar to the SVM,
but is capable of providing a probabilistic interpretation to
predictions. For target values $y \in\{0,1\}$ (remapping the input
training labels from $\{-1,+1\}$ is trivial), then predictions are of
the form,

\begin{align}
\hat{y}\left(\vec{x} \right)=\sum_{i=1}^n
w_ik\left(\vec{x},\vec{x}_i\right) + b
\end{align}
\noindent
Where the $b$ is essentially analogous to the bias parameter in the
SVM. In this case, the $w_i$ are constructed through an iterative
optimization process where there is an initial assumption,

\begin{align}
\mathbb{P}\left(\vec{w}\vert \vec{\alpha}\right) = \prod_{i=1}^m
\mathcal{N}\left(w_i\vert 0, \alpha_i^{-1} \right)
\end{align}

\pagebreak[4]
\noindent
and the $\alpha_i$ are precision parameters corresponding to the
$w_i$. Rather than taking the sign of the linear function, as was the
case in the SVM situation, a
classification decision is constructed by considering a logistic
sigmoid function as follows:

\begin{align}
\mathbb{P}\left(y\vert \vec{x} \right) \equiv \sigma\left(\sum_{i=1}^n
w_ik\left(\vec{x},\vec{x}_i\right) + b\right)
\end{align}
\noindent
Such that,
\begin{align}
\sigma\left(\theta\right) = \frac{1}{1+e^{-\theta}}
\end{align}
\noindent
In particular, if $\mathbb{P}\left(y\vert \vec{x} \right) >
\frac{4}{5}$ then a classification decision of $+1$ is returned,
whereas if the converse is true, then a classification decision of
$-1$ is returned (after remapping into the original target-space)
\cite{tipping,bishop}. The threshold of $\frac{4}{5}$ was selected by
experimentation as delivering strong predictive results. This
selection can be justified intuitively in some sense by the
observation that high confidence is preferable in stock prediction,
and that one is more likely to invest correctly when the probability
of class membership is 0.80 rather than 0.50. The RVM is then capable
of providing an interpretation that a given stock input has a
particular class membership probability above the 0.80 threshold,
which the user can incorporate into further analyses.  

Summarily speaking, the RVM is an advantageous algorithm to use
due to its probabilistic properties, and the value of the sigmoid
function represents a degree of confidence that a given feature vector
belongs to the class of stocks with positive returns from an initial
quarter to a subsequent quarter.

\subsection{$k$-Nearest Neighbors Ensemble}

A disadvantage of parametric kernel methods in the style of the SVM
and the RVM is that, for poor choices of the parameter $\gamma$, the
learned model will fail to detect relevant patterns within the data
and the model will demonstrate very lacking performance. In the
context of financial learning, a bad $\gamma$ may result in the
learner labeling \emph{every} test case as a stock giving positive (or
negative) returns, even though in actuality this can scarcely be true.  

This issue can be addressed by implementing a classifier that
estimates $\mathbb{P}\left(\mathcal{C} = \pm 1 \vert \vec{x} \right)$
across the feature space using a naive Euclidean Distance metric. This
classification algorithm is known as a $k$-Nearest Neighbors ($k$-NN)
method. In particular,
after applying Bayes' Rule \cite{bishop} we arrive at a class
posterior probability distribution,

\begin{align}
\mathbb{P}\left(\mathcal{C} = \pm 1 \vert \vec{x} \right) =
\frac{\mathbb{P}\left(\vec{x}\vert \mathcal{C} \right)\mathbb{P}\left(
  \mathcal{C}\right)}{\mathbb{P}\left(\vec{x} \right)}=\frac{K_{\mathcal{C}}}{K}
\end{align}
\noindent
Where $K$ is the number of closest (Euclidean) neighbors to consider and $K_{\mathcal{C}}$ is the number of points of class $\mathcal{C}$ that are within the
set of $K$ closest neighbors to the vector $\vec{x}$. Thus, a
classification decision may determined by maximizing the quantity
$\frac{K_{\mathcal{C}}}{K}$ with respect to the class label. 

Empirically speaking, it is the case that $k$-NN classifiers are not
strong classifiers in the sense that they are capable of delivering
high accuracy as a single unit. Therefore, we train a committee of one hundred
weak $k$-NN classifiers using a subset of the training examples for
each constituent learner. The parameter $k$ is selected through a 10-fold
cross-validation process where the $k$ minimizing the average error is
used to train a model for the larger ensemble.

\subsection{Boosting Classifier Performance}

Financial time-series forecasting suffers from something of a bad
reputation, being based on noisy data for which it is often
intractable to learn an effective model. Techniques exist in machine learning to
relieve this problem and within the scope of this work we adopted the
method of boosting for improving a basis set of ``weak
learners'' to generate a more capable predictor. 

\vspace{3mm}
\begin{center}
\line(1,0){250}
\end{center}
\begin{algorithm}[H]
 \SetAlgoLined
 \KwData{Let $\mathcal{D}_{\text{train}}$ denote a training data set
   and $\mathcal{D}$ denote the testing data set. }
 \KwResult{An ensemble model augmented by boosting and an evaluation
   of the model on the testing data set.}
 $\mathbb{M} =  \left\{\text{Random Forest}_{\text{model}},\text{SVM}_{\text{model}},\text{RVM}_{\text{model}},k\text{-NN}_{\text{model}} \right\}$\;
 \For{$\hat{y}_i(\vec{x}) \in \mathbb{M}$}{
   Learn the model $\hat{y}_i(\vec{x})$ on $\mathcal{D}_{\text{train}}$ by
   performing the optimization,
   \begin{align}
     J_i =\text{min}~ \sum_{j=1}^{\vert
       \mathcal{D}_{\text{train}}\vert} \mathbb{I}\left(\hat{y}_i\left(\vec{x}^{(j)}\right)\neq y^{(j)} \right)
   \end{align}
   Where $\mathbb{I}(*)$ is the indicator function. Perform the evaluations,
   \begin{align}
     \epsilon_i = \frac{J_i}{\sum_{j=1}^{\vert
         \mathbb{D}_{\text{train}} \vert} w_i^{(j)}}\\
     \alpha_i = \log \left[ \frac{1-\epsilon_i }{\epsilon_i} \right]
   \end{align}
 }
 Using the ensemble model that has been augmented by boosting, use a
 final decision criterion of the form,
 \begin{align}
   \hat{y}_{\text{Boost}} = \text{sign} \left[ \sum_{i=1}^{\vert
       \mathbb{M} \vert} \alpha_i \hat{y}_i(\vec{x}) \right]
 \end{align}
 And for all $\vec{x} \in \mathcal{D}$ this gives the prediction of
 the boosted ensemble on the testing data.
 \vspace{3mm}
 \caption{Implementation of boosting in the context of the financial
   ensemble model. Notice that this algorithm is similar to AdaBoost,
   yet is forced to accept a different objective function.}
\end{algorithm}
\begin{center}
\line(1,0){250}
\end{center}

The essential idea behind the boosting algorithm is that if the first
classifier performs strongly and accurately predicts the stock return
outcome, then it is less important for subsequent classification
algorithms to also make these same correct predictions, and therefore
their impact on the prediction task becomes less meaningful and less
necessary. A direction for future research in this area would be in
application to the discovery of methodologies for incorporating a
\emph{weighted} objective function that more finely balances the
constituent models (consider for instance an augmentation of Adaptive
Boosting). 

In some sense of the word, this implementation of boosting is a ``hacky'' solution,
relying chiefly on inspiration from a grossly simplified version of AdaBoost and on an
empirical evaluation of simply \emph{what works well}. Nonetheless, it was
determined experimentally that this algorithm learns weighting
coefficients of comparable accuracy to an exhausting grid search of
parameters on the testing data. Therefore, though the algorithm lacks
a theoretical framework, its efficacy in practice justifies its presence.

\section{Meta-Parameter Selection for Parametric Learners}

In the ensemble we have elected the use of the SVM and RVM parametric
learners. We conduct an extensive series of cross-validation
evaluations to select the best-performing value of $\gamma$ in the
kernel function. In particular, we partition the training data
$\mathcal{D}_{\text{train}}$ into five randomly chosen subsets. For
every candidate value of $\gamma$, a model is learned by excluding one
of the partitions from the training. This excluded division is then
used to test the learned model and an error rate is recorded. The
candidate value of $\gamma$ that performs best on average is chosen to
be the best meta-parameter and is used to formulate the full model. 


\begin{table}[H]
\centering
\begin{tabular}{|l|l|p{9cm}|}
\hline
\multicolumn{2}{c}{Explanatory Variable} \\
\cline{1-2}
\emph{Variable Index} & \emph{Variable Name} & \emph{Description}
\cite{data}\\
\hline
1 & ACTQ & Current assets (total)\\\hline
2 & CHEQ & Cash and short-term investments\\\hline
3 & DLCQ & Debt in current liabilities\\\hline
4 & DLTTQ & Long-term debt (total)\\\hline
5 & EPSPXQ & Earnings per share (basic and excluding extraordinary items)\\\hline
6 & EPSX12 & Earnings per share in 12 months (basic and excluding
extraordinary items)\\\hline
7 & ICAPTQ & Quarterly invested capital (total)\\\hline
8 & LCTQ & Current liabilities (total) \\\hline
9 & LTQ & Liabilities (total)\\\hline
10 & NIQ & Net income (loss)\\\hline
11 & OEPS12 & Earnings per share from operations (12 months
moving)\\\hline
12 & OIADPQ & Operating income after depreciation (quarterly)\\\hline
13 & REVTQ & Revenue (total and quarterly)\\\hline
14 & SPCE12 & S\&P core earnings (12MM)\\\hline
15 & SPCEQ & S\&P core earnings\\\hline
16 & WCAPQ & Working capital (balance sheet)\\\hline
17 & XOPRQ & Operating expense (total and quarterly)\\\hline
18 & CAPXY & Capital expenditures\\\hline
19 & EPSFIY & Earnings per share (diluted and including extraordinary items)\\\hline
20 & IVCHY & Increase in investments\\\hline
21 & REVTY & Revenue (total and yearly)\\\hline
22 & SPCEDY & S\&P core earnings EPS diluted\\\hline
23 & SPCEEPSPY & S\&P core earnings EPS basic (preliminary)\\\hline
24 & SPCEPY & S\&P core earnings (preliminary)\\\hline
25 & XOPRY & Operating expense (total and yearly)\\\hline
26 & CSHTRQ & Common shares traded (quarterly)\\\hline
27 & MKVALTQ & Market value (total)\\\hline
28 & PRCCQ & Price close (quarter)\\\hline
29 & PRCHQ & Price high (quarter)\\\hline
30 & PRCLQ & Price low (quarter)\\\hline
\end{tabular}
\caption{The varying variables used for stock return prediction in the
context of this work. Every variable is given a index for the purposes
of identification within graphics where full text would have been
cumbersome. A short description of what each variable captures in the
financial analysis is also
provided to the right of each variable.}
\label{tab:datatable}
\end{table}

The partial advantage of the RVM over its SVM counterpart is the lack
of need to tune the so-called ``slackness'' parameter, $C$, that represents,
intuitively, the extent to which the SVM is permitted to violate the
presumed linear separability of the data (often after it has been
remapped to the feature space by the kernel). This parameter was
investigated thoroughly by Huerta, who determined that the assignment
$C = 2$ was most often preferred in a cross-validated process
\cite{huerta}. Following this previous work, we do not increase the
complexity of the model training algorithm by fine-tuning the
slackness parameter. Instead, $C=2$ is chosen a priori. 


\section{Data Description and Prefiltrations}

The data used in this work were obtained from the Wharton Research
Data Services (WRDS), which provided access to CRSP and Compustat
databases. Data are considered from the years 2006 through 2012. This interval of financial history is of particular relevance
because it reflects a period in time that was characterized by
financial collapse and, to some extent, eventual recovery. This range
of situations presents the learned model with a dynamic set of
circumstances in which it may be applied. 

We provide a table (see Table \ref{tab:datatable}) of the explanatory variables incorporated into the analysis of this work, where
each explanatory variable is assigned both a index for reference and a
brief description from the Compustat database. 

\subsection{GICS Partitioning and Motivation}

A major element of this work is to examine the effectiveness of the
ensemble model on financial data from the various Global Industry
Classification Standard (GICS) sectors. The GICS sectors are: Energy, Materials,
Industrials, Consumer Discretionary, Consumer Staples,
Health Care, Financials, Information Technology,
Telecommunication Services, and Utilities. As discovered by Huerta, there exist
certain industry sectors (in particular, the Telecommunications and
Utilities sectors) that do not possess a ``critical mass'' of data
vectors to effectively divide into training and testing
categories. In particular, the numerical calculations required in
formulating the RVM model tend to lack the robustness for these
low-density training sets and this condition prevents the model from terminating
successfully. Therefore, these sectors are not considered in terms of
deploying the model. 

Partitioning the financial data according to the GICS sector is
motivated first by an interest to observe differences is learned
explanatory variables between industries. In other words, it is
believed that technical variables have explanatory powers that vary
across industry sectors, and that by selectively choosing these
factors for each sector, the predictive capabilities of model will be
generally improved. The partitioning is further motivated by an
interest in improving the deployability of the model. The algorithmic
complexity of these models increases considerably as additional
vectors are added to the training set; thus, to improve the time to
termination of the model, it is beneficial to decrease the number of
points in the training set, which is conveniently and justifiably accomplished by training
a model for every GICS sector. 

\subsection{Supervised Feature Selection with a Relief Algorithm}

A key attribute of this project was the automation of feature
selection within the varying industry sectors. This approach allows
the ensemble learner to evaluate for itself the relevance of
explanatory variables to prediction, and does not rely on human
specification or on prior domain-expert review. In a sense, this
automated feature estimation constitutes a ``purer'' learning
methodology than does pre-specification. 

A Relief algorithm is chosen for the purpose of variable selection because they have
demonstrated an empirical capability to detect dependencies within the
data from both regression and classification problems. In particular,
we implemented the Relief-F algorithm to identify critical features
across industries. Relief-F is advantageous for selecting informative
explanatory variables because it is comparatively less sensitive to
stochastic input data than other Relief algorithms
\cite{reliefadvantage}. The essential idea behind the Relief-F
algorithm is to select at random examples in the training data,
observe their $k$ nearest neighbors, and assign a weighting to each
explanatory variable according to how well it distinguishes the
example from nearby data points of separate class \cite{relieff}. In particular, the
Relief-F algorithm estimates the probabilities and arrives at a
weighting for each feature $i$,

\begin{align}
w_i = \mathbb{P}\left(f \neq i \vert \mathcal{C} \neq
  \mathcal{C}_{\vec{x}} \right) - \mathbb{P}\left(f \neq i \vert \mathcal{C} =
  \mathcal{C}_{\vec{x}} \right)
\end{align}

\noindent
This construction is intuitively thought of as the probability that
the feature $f$ takes a value different from the value $i$ given that
the classes are not identical, minus the probability that $f$ assumes
a different value yet the classes are the same \cite{relieff}. 

The algorithm for the Relief-F is written explicitly as follows,

\vspace{3mm}
\begin{center}
\line(1,0){250}
\end{center}
\begin{algorithm}[H]
 \SetAlgoLined
 \KwData{For all vectors in the data set under consideration,
   $\mathcal{D}$ assign attribute values and class labels. Let
   $\mathcal{D}$ be an $m$-by-$n$ matrix.}
 \KwResult{A weighting vector $\vec{w}$ whose components are in the
   interval $\left[-1,+1 \right]$, where increased positivity
   indicates increased relevance to classification.}
 $\vec{w}_i = 0~\forall ~i \in \{1,\ldots,n \}$\;
 \For{$i \in \{1,\ldots,k \}$, for a $k$ nearest neighbors algorithm}
 {
   Select a random training instance $x_i$\;
   $H = \text{The number of nearest neighbors with identical class}$\;
   $M = \text{The number of nearest neighbors with separate class}$\;
   \For{$j \in \{1,\ldots, n \}$} {
     $h_j =\text{Proportion of nearest neighbors of identical class
       whose}~j^{\text{th}}~\text{entry matched}~x_{ij}$\;
     $m_j =\text{Proportion of nearest neighbors of separate class
       whose}~j^{\text{th}}~\text{entry matched}~x_{ij}$\;
     $w_j = w_j + \frac{H_j - m_j}{k}$\;
   }
 }
 \vspace{3mm}
 \caption{Implementation of the Relief-F algorithm for financial
   ensemble feature selection. This implementation is biased toward
   the $k$-Nearest Neighbor approach, however empirical evidence has
   suggested that Relief-F is a powerful algorithm for determining
   relevant features in classification problems.}
\end{algorithm}
\begin{center}
\line(1,0){250}
\end{center}

\section{Results and Analysis of the Ensemble}

Presented here are a series of empirical experiments developed to
demonstrate concurrently both the strengths and weaknesses of a
learning algorithm of this form. In particular, we demonstrate through
empirical evidence that this learning algorithm has an ability to form
effective predictions in cases where at least one of the constituent
algorithms performs weakly in terms of stock price return
prediction. However, some results indicate that substantial
overfitting can occur in instances where too much data is presented to
the model. Furthermore, the boosting procedure tends to learn only to
imitate a constituent learner's solution in cases where all members of
the ensemble perform strongly. 

\begin{figure}[H]
        \centering
        \begin{subfigure}[b]{0.3\textwidth}
                \centering
                \includegraphics[width=\textwidth]{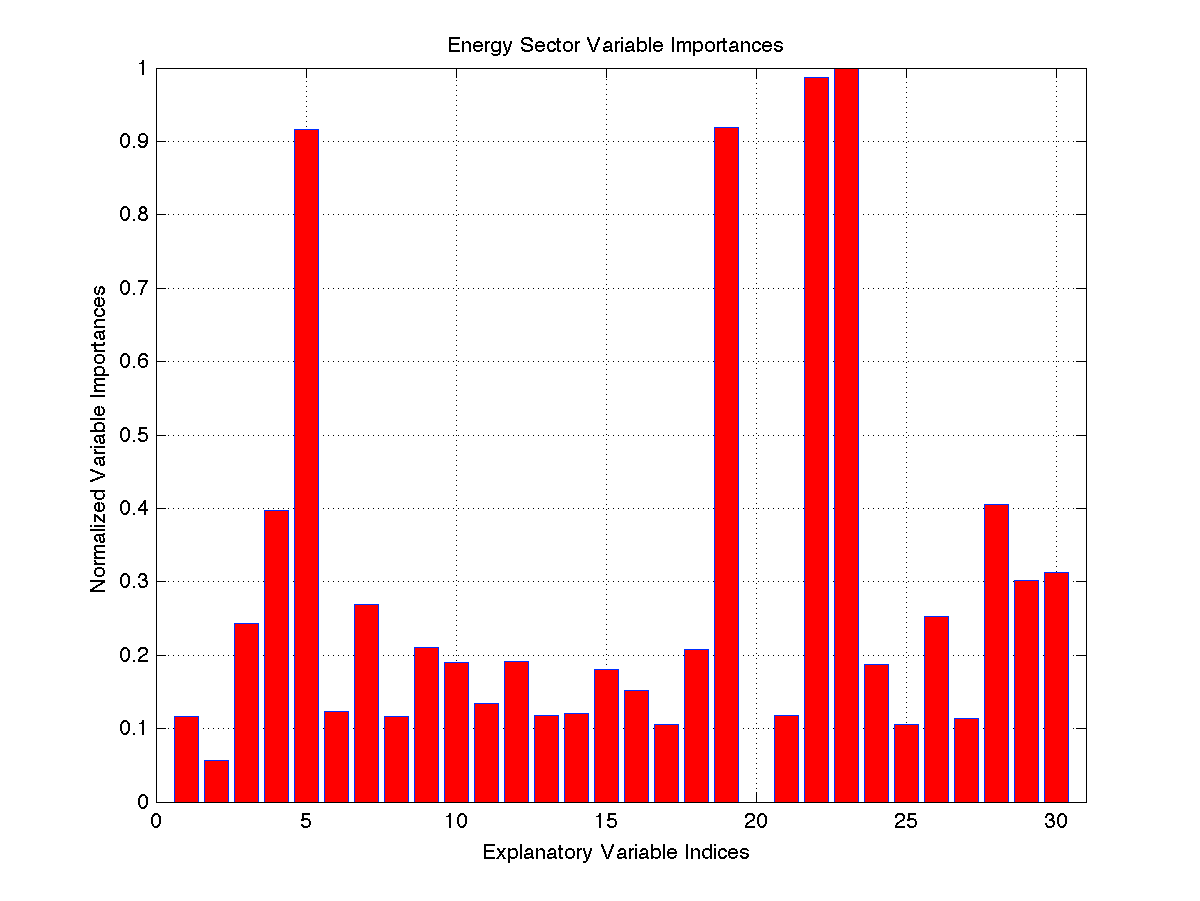}
                \caption{Energy: S\&P core earnings EPS basic (prelim.)}
        \end{subfigure}%
        ~ 
        \begin{subfigure}[b]{0.3\textwidth}
                \centering
                \includegraphics[width=\textwidth]{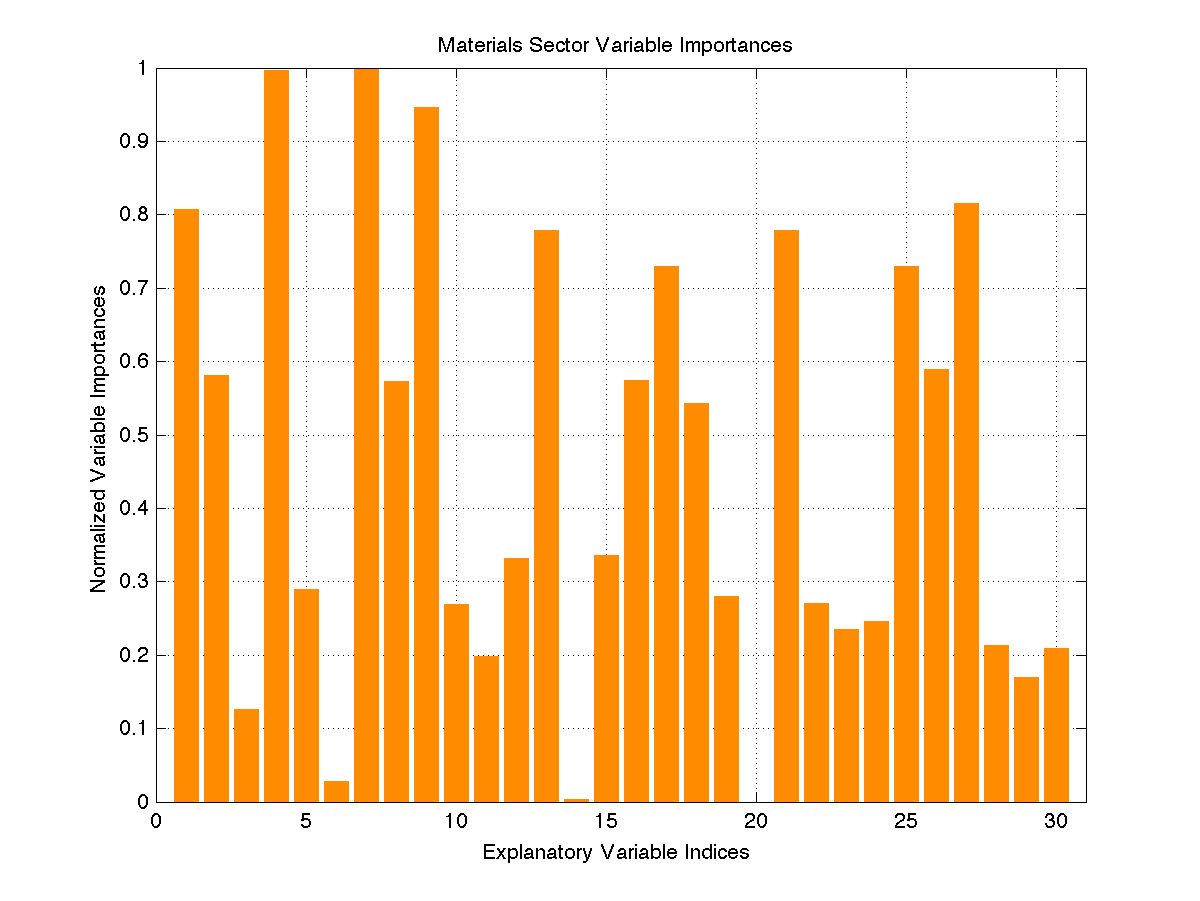}
                \caption{Materials: quarterly invested capital (total)}
        \end{subfigure}
        ~
       \begin{subfigure}[b]{0.3\textwidth}
                \centering
                \includegraphics[width=\textwidth]{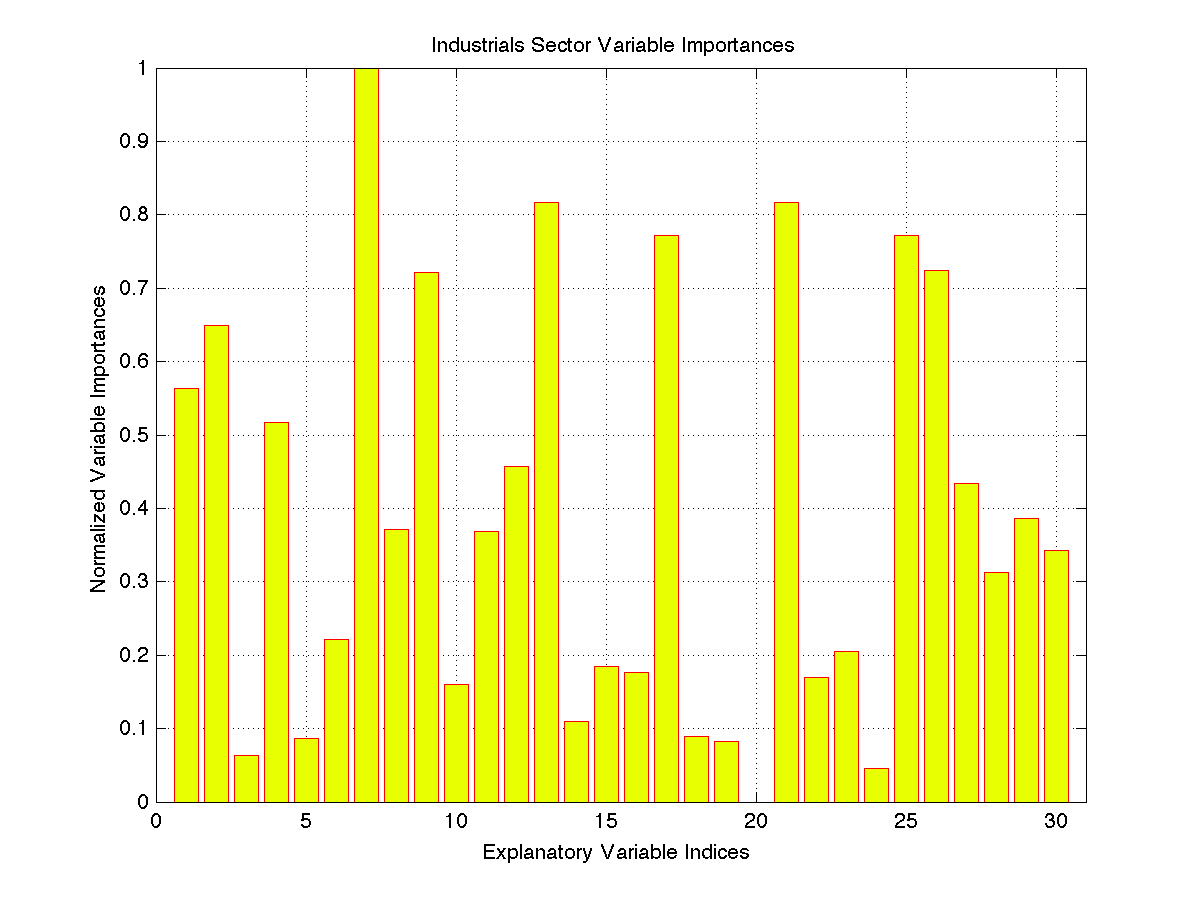}
                \caption{Industrials: quarterly invested capital (total)}
        \end{subfigure}

        \begin{subfigure}[b]{0.3\textwidth}
                \centering
                \includegraphics[width=\textwidth]{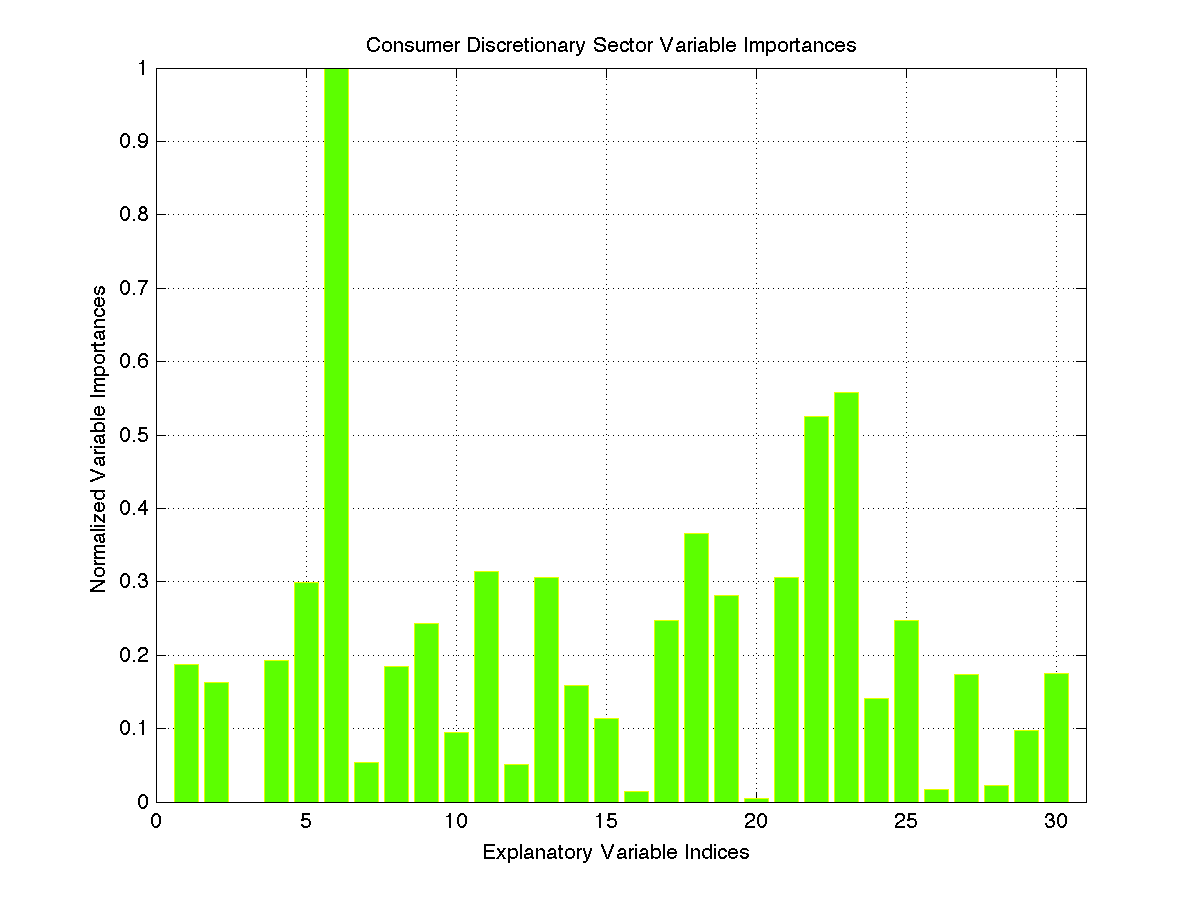}
                \caption{Consumer Discretionary: earnings per share in
                12 months}
        \end{subfigure}
        ~
        \begin{subfigure}[b]{0.3\textwidth}
                \centering
                \includegraphics[width=\textwidth]{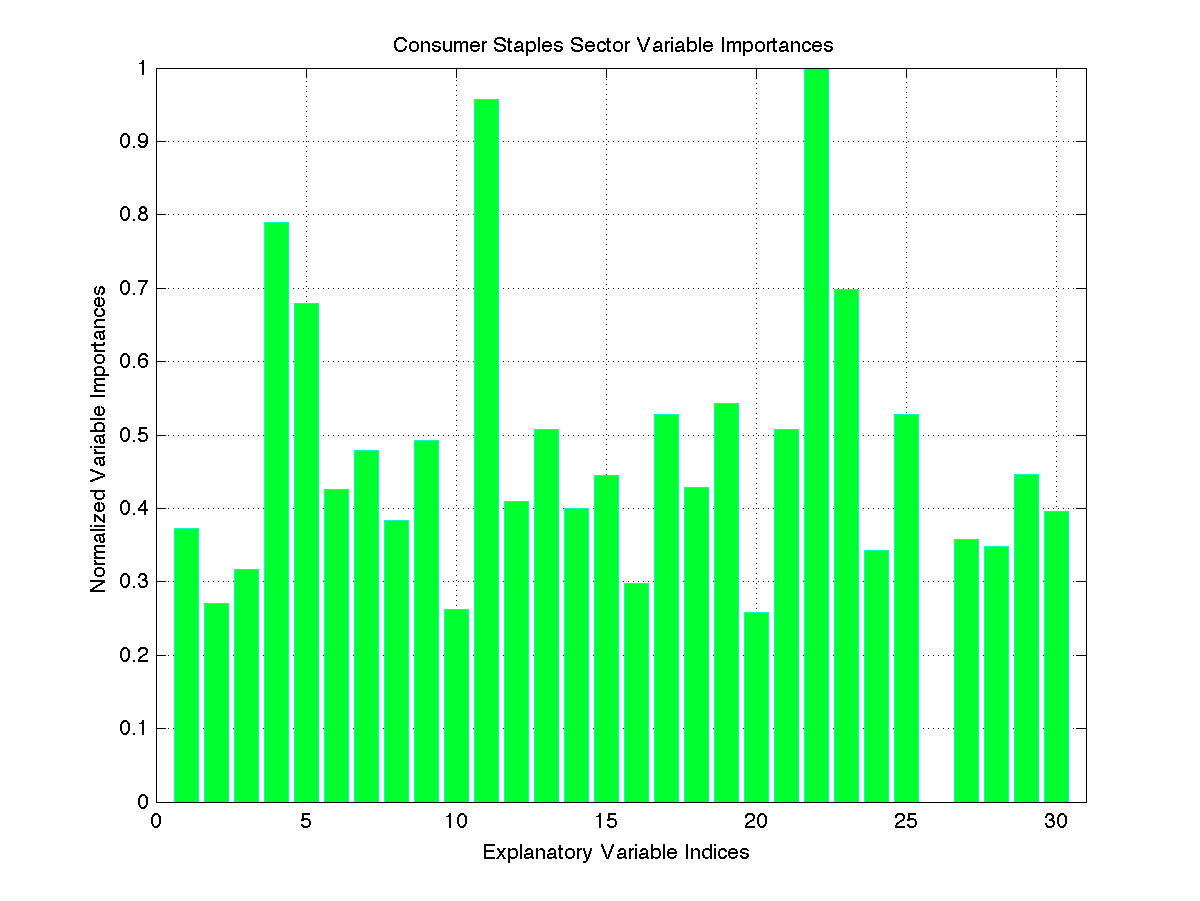}
                \caption{Consumer Staples: S\&P core earnings EPS diluted}
        \end{subfigure}
        ~
        \begin{subfigure}[b]{0.3\textwidth}
                \centering
                \includegraphics[width=\textwidth]{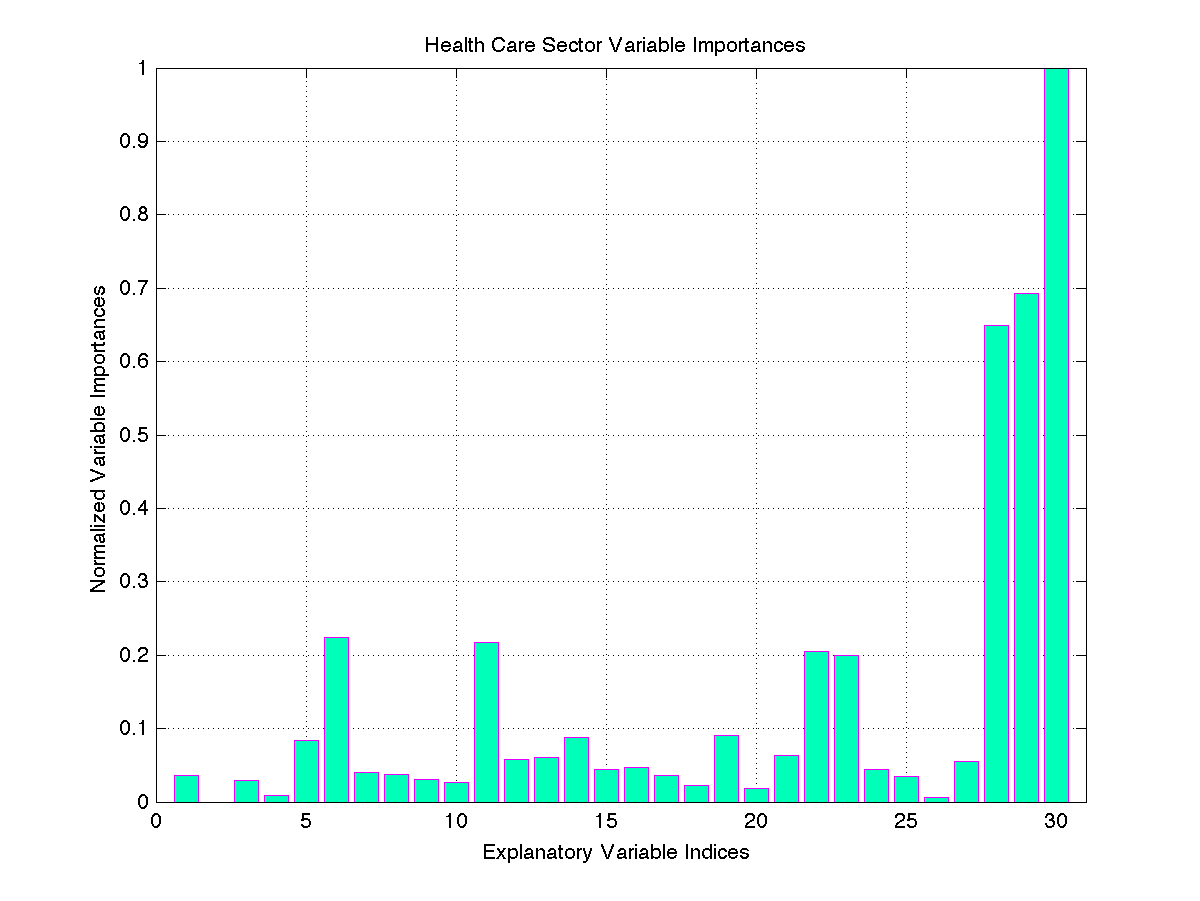}
                \caption{Health Care: price low (quarterly)}
        \end{subfigure}
        
        \begin{subfigure}[b]{0.3\textwidth}
                \centering
                \includegraphics[width=\textwidth]{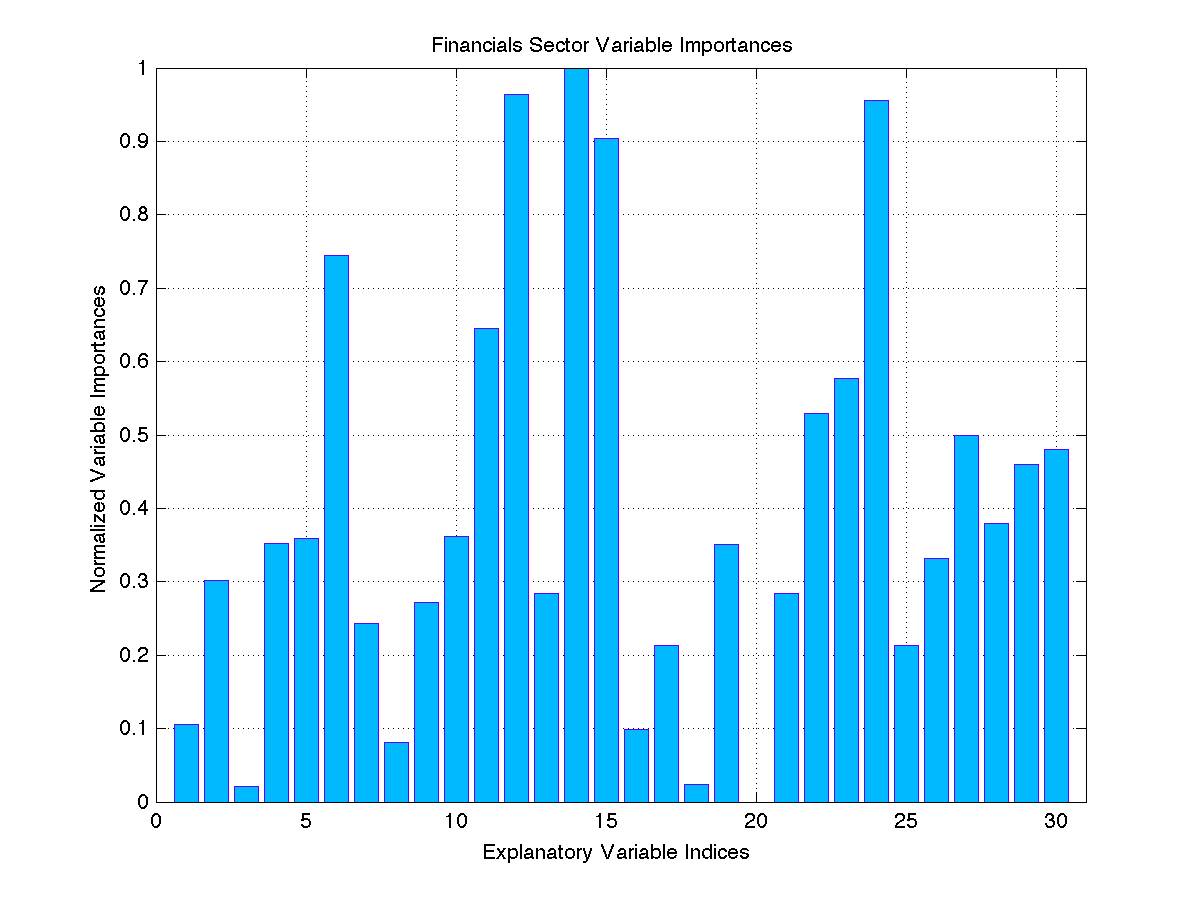}
                \caption{Financials: S\&P core earnings (12MM)}
        \end{subfigure}
        ~
        \begin{subfigure}[b]{0.3\textwidth}
                \centering
                \includegraphics[width=\textwidth]{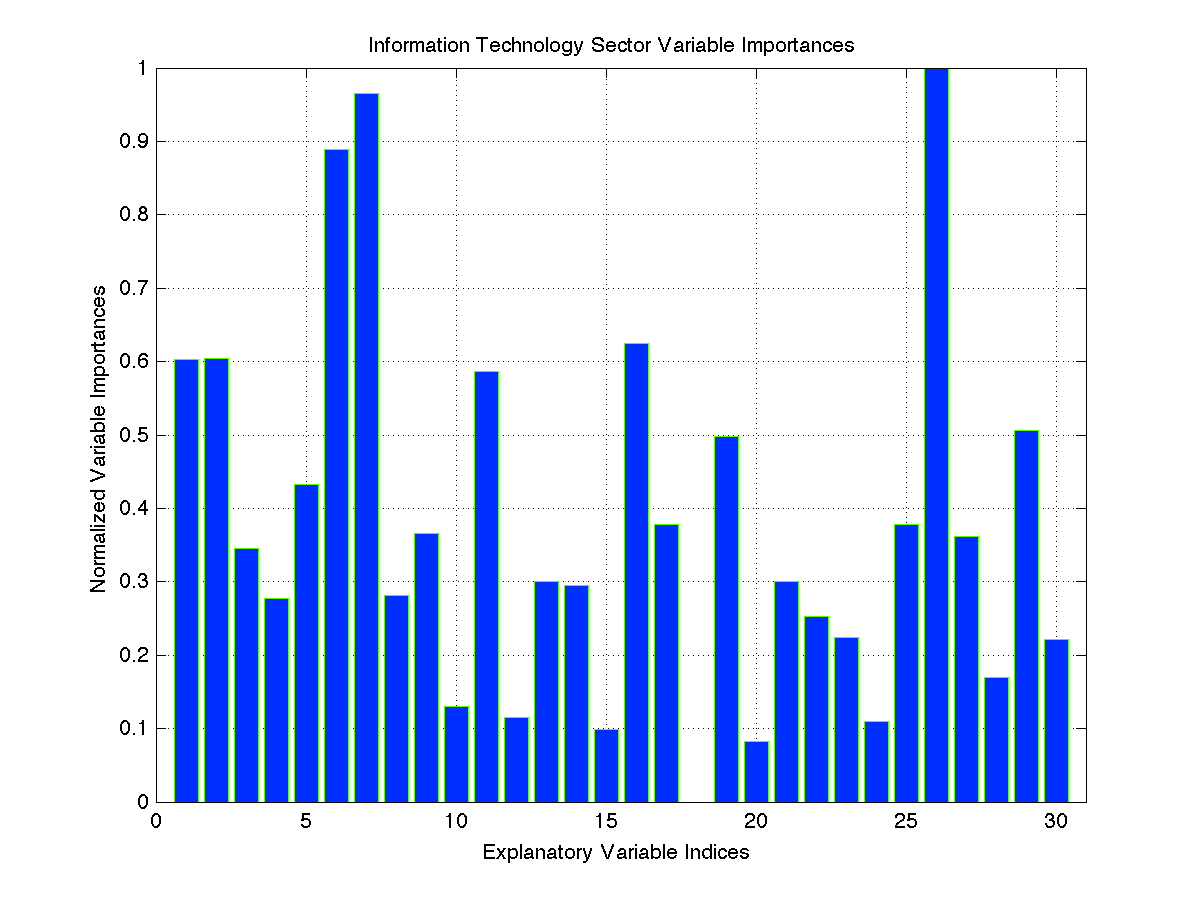}
                \caption{Information Technology: common shares traded}
        \end{subfigure}
        ~
        \begin{subfigure}[b]{0.3\textwidth}
                \centering
                \includegraphics[width=\textwidth]{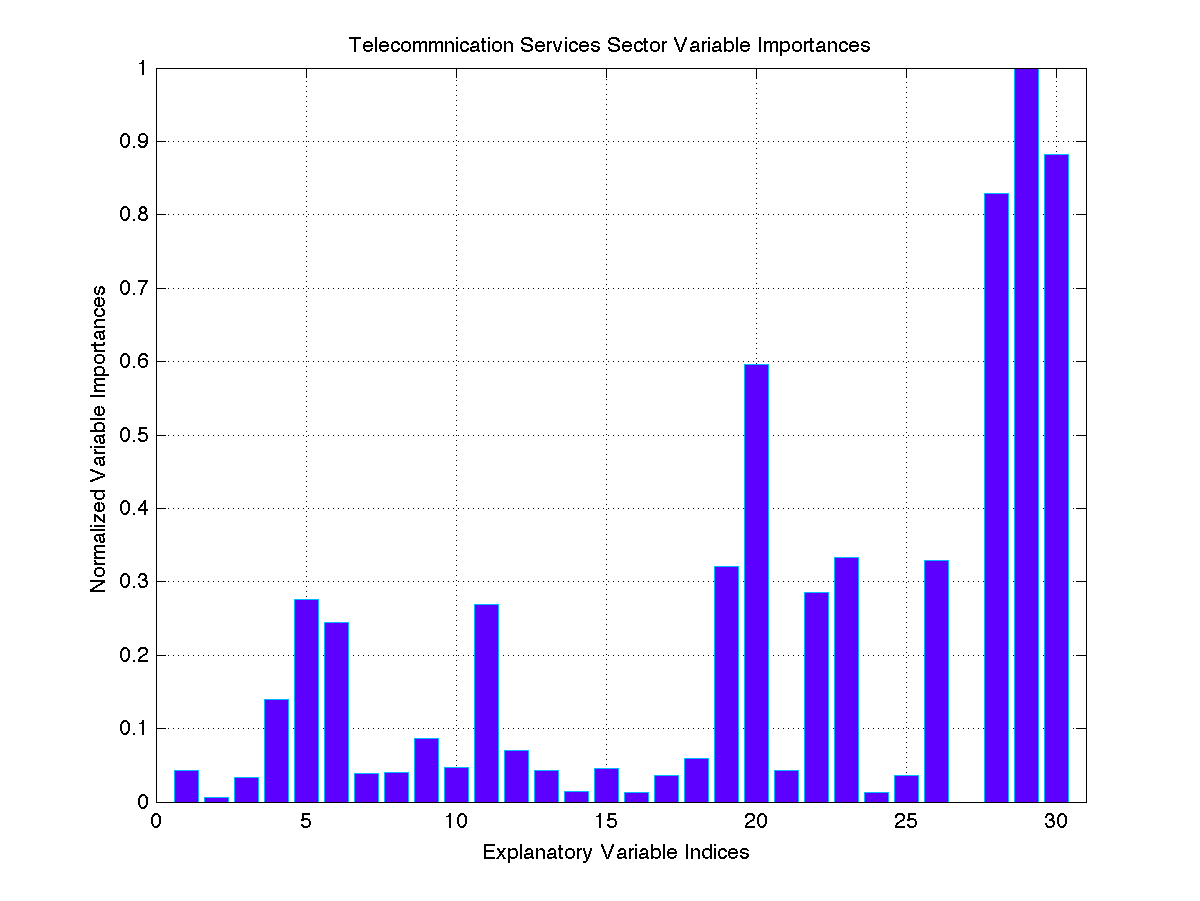}
                \caption{Telecommunication Services: price high (quarterly)}
        \end{subfigure}
        
        \begin{subfigure}[b]{0.3\textwidth}
                \centering
                \includegraphics[width=\textwidth]{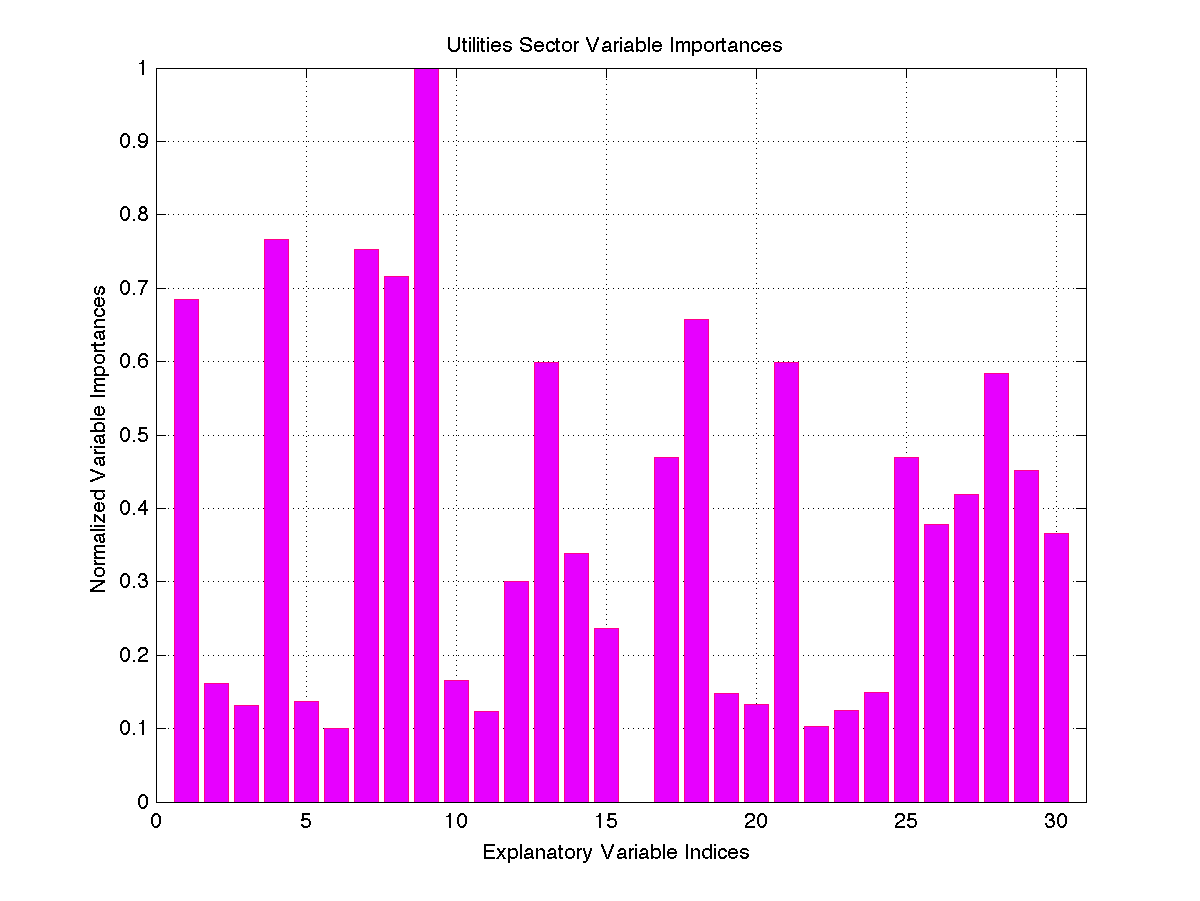}
                \caption{Utilities:\\ liabilities (total)}
        \end{subfigure}
        ~
        \begin{subfigure}[b]{0.3\textwidth}
                \centering
                \includegraphics[width=\textwidth]{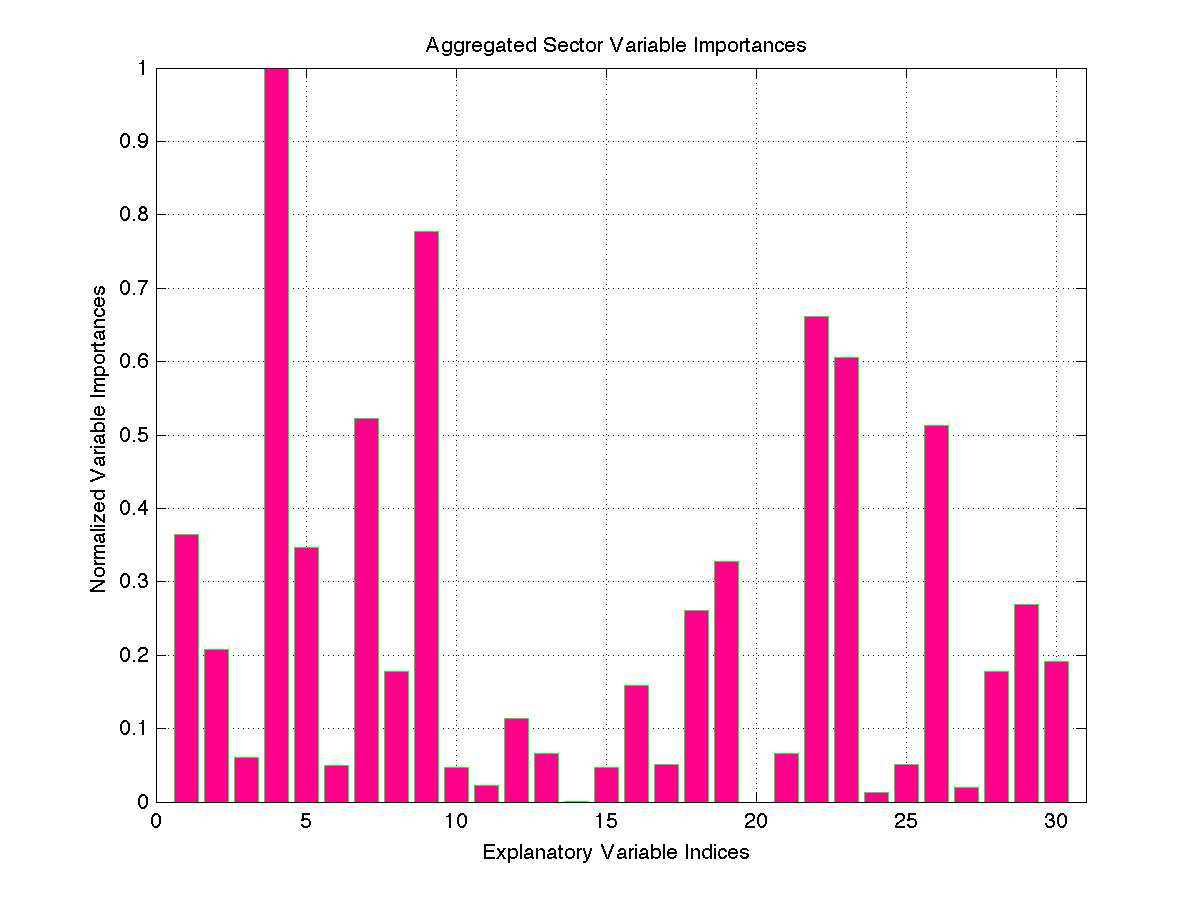}
                \caption{Aggregated: long-term debt (total)}
        \end{subfigure}
\caption[8pt]{Industry-specific variable importances for all GICS in
  the interval from the first quarter to the second quarter of
  2009. The sector is stated and then the dominant training variable
  according to Relief-F is provided as well.\\

Here the weights are \emph{linearly} normalized to be in the
interval $\left[0,1 \right]$, where higher values indicate features of
greater relevance for a particular industry. The period under
consideration was selected arbitrarily and for the purposes of
visualization and example.} 
\label{fig:bars}
\end{figure}

As discussed in previous section, identifying important,
industry-specific predictors was a major consideration in this
project. Figure \ref{fig:bars} demonstrates the relevances of particular
explanatory variables across all of the GICS. For the purposes of
visualization, the variable importances returned by the Refief-F
algorithm have been normalized to the interval $[0,1]$ by a linear
scaling procedure. In particular, low values toward zero indicate weak explanatory
power, while high values are characteristic of highly predictive
variables for that sector. The visualization of these results indeed
reveal that the importances of predictors do vary across industry
sectors in terms of their efficacy in predicting stock price
returns. See Table \ref{tab:datatable} for a summary of the explanatory variables used in this
analysis and their corresponding indices. 


From the discussion of the constituent algorithms incorporated into
the ensemble, it is apparent that cross-validation serves a vital role
in formulating an effective model for stock return prediction. For
both the support vector machine and the relevance vector machine are
evaluated for a choice of $\gamma$ from the discrete set
$\{\frac{1}{2},1,2,4 \}$, where these values are selected for
consistency with Huerta. Further, these choices of parameter were
evaluated in terms of efficacy by five-fold cross-validation, where an
average of errors is used to determine the best choice. In the case of
the SVM, the cross-validation procedure is consistent with Huerta as
the assignment $\gamma =2$ is the most common selection. 



It was surprising to observe that the RVM does not follow this choice and
instead most often chooses a $\gamma$ unique to industry, though the
functional form of the model is identical to the SVM. This phenomenon
can be explained to an extent by the observation that, in practice,
the relevance and support vectors selected by either algorithm for
prediction are rarely the same, and more often the support vectors are
near to decision boundaries, whereas relevance vectors are more
antithetical in nature \cite{tipping}. It therefore makes intuitive
sense that a different distance augmentation would be preferable for
the classification task. This result of industry-related impacts on
parameter selection is summarized for the year 2009 in Table
\ref{tab:cross}.

\begin{table}[H]
\centering
\begin{tabular}{|l|l|l|}
\hline
\multicolumn{2}{c}{Cross Validation across Industries for RVM} \\
\cline{1-2}
\emph{GICS Name} & \emph{Hyperparameter Selection} & \emph{Frequency
  of Selection}\\
\hline
Energy & 4.0 & $\sim$50\%\\\hline
Materials & 0.5 & $\sim$80\%\\\hline
Industrials & 0.5 & $\sim$95\%\\\hline
Consumer Discretionary & 4.0 & $\sim$65\%\\\hline
Consumer Staples & 1.0 & $\sim$80\%\\\hline
Health Care & 2.0 & $\sim$80\%\\\hline
Financials & 0.5 & $\sim$95\%\\\hline
Information Technology &1.0 & $\sim$75\%\\\hline
\end{tabular}
\caption{The cross-validated selections for the $\gamma$ parameter
  across industry sectors in the period from the first quarter of 2008
to the second quarter. Presented here are the most frequently occurring
selections for the parameter and the frequency with which it occurred
in 50 runs of the ensemble.}
\label{tab:cross}
\end{table}

\begin{figure}[h]
\begin{subfigure}[b]{0.5\textwidth}
\centering
\includegraphics[width=\textwidth]{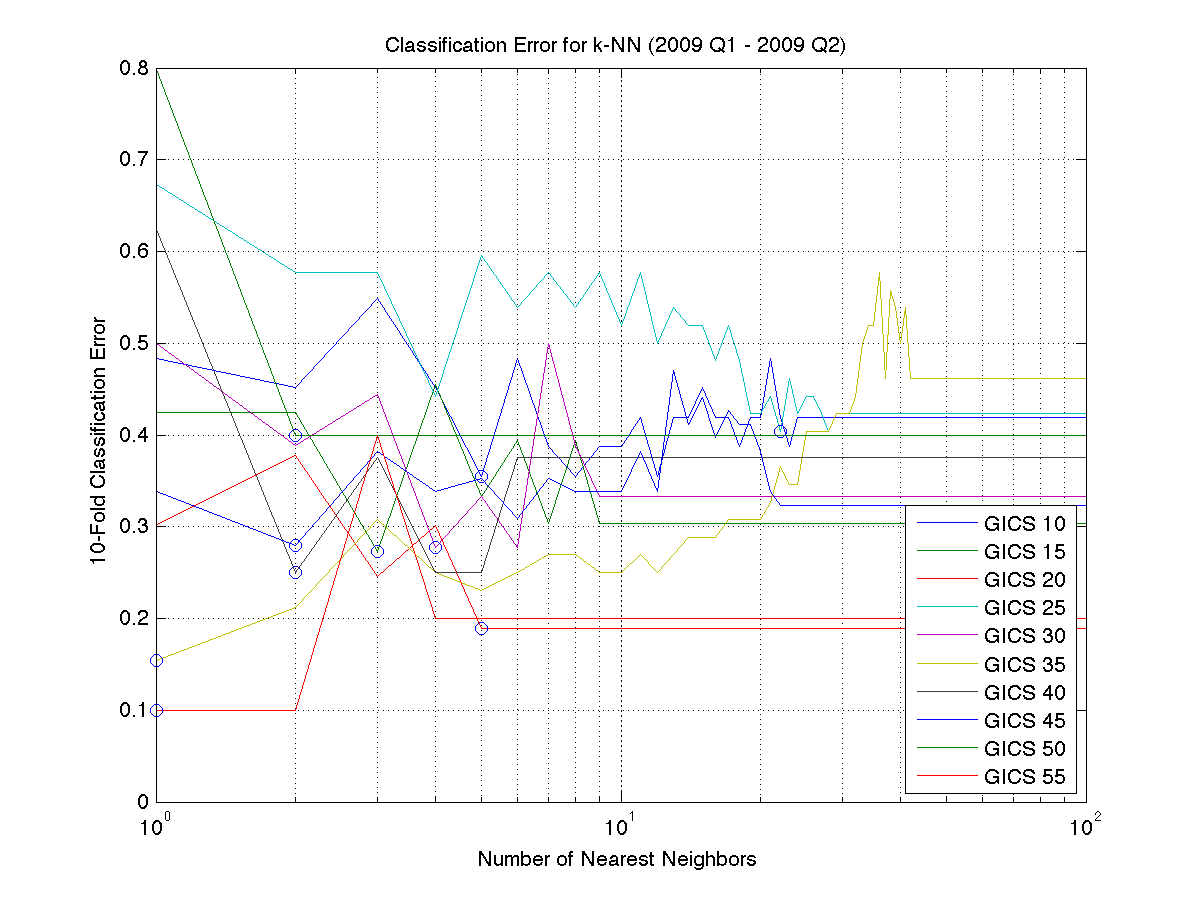}
\captionsetup{font=scriptsize}
\caption[8pt]{A survey of all industry sectors and the $k$-Nearest
  Neighbor's effectiveness for varying $k$. The first instances of a
  minimum are indicated by a blue circle.}
\label{fig:knn}
\end{subfigure}
~
\begin{subfigure}[b]{0.5\textwidth}
\centering
\includegraphics[width=\textwidth]{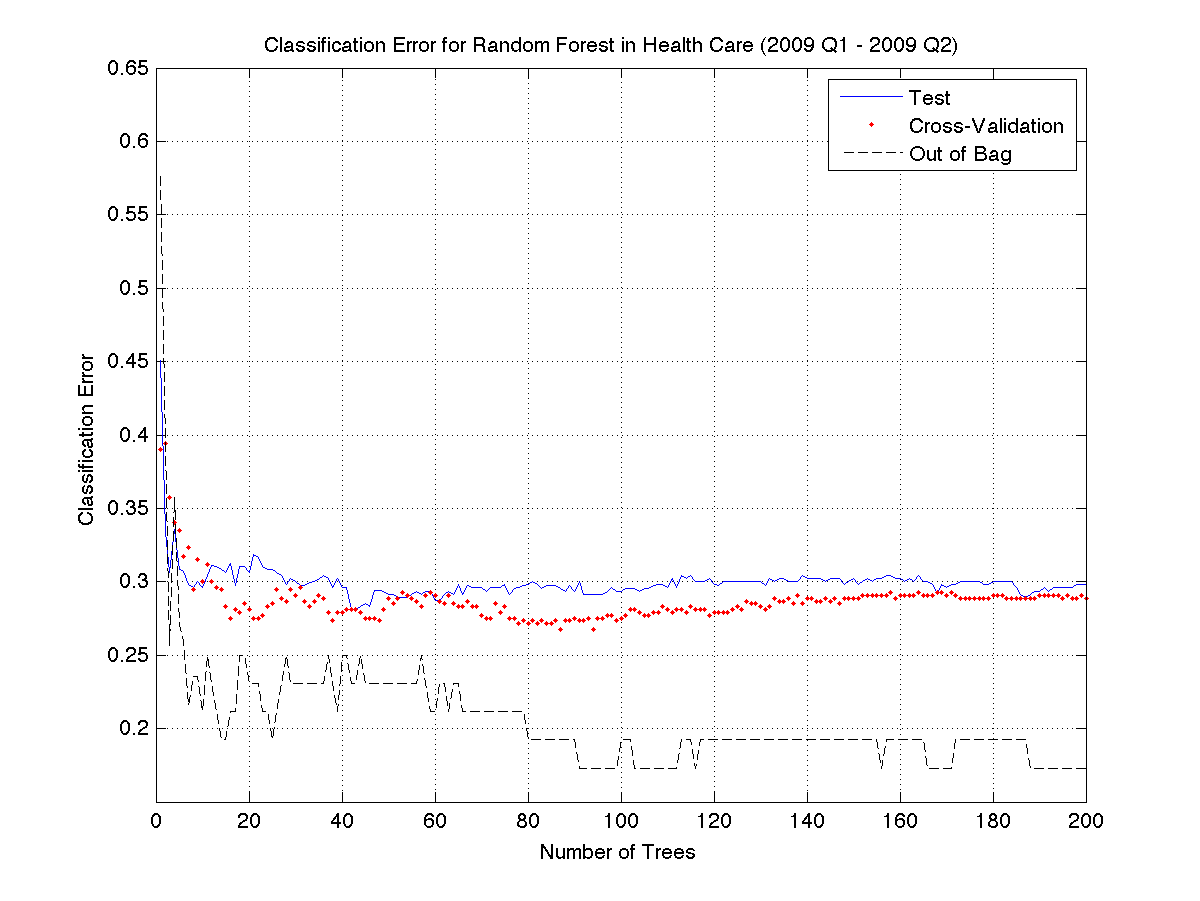}
\captionsetup{font=scriptsize}
\caption[8pt]{Comparison of classification error metrics in the
  Random Forest algorithm. Presented are the test error, the
  cross-validated error, and the Out-of-Bag estimate of the error.}
\label{fig:rf}
\end{subfigure}

\caption{Notice in the $k$-Nearest Neighbors visualization that these results are presented in a logarithmic scale
in $k$. This is done simply in the interest of a better visualization,
and is not meant to suggest that there is an apparent pattern in the
error when plotted against a logarithmic $k$. Because these
cross-validated results required the partitioning of the training data
into tenths, it is entirely conceivable that, for particular $k$, $k > \frac{9}{10}
\vert \mathcal{D}_{\text{train}}\vert$. In this case, $k$ was reduced
to the cardinality of the nine-tenths partitioning of the training
data.}
\end{figure}

In constructing the $k$-Nearest Neighbor predictor it was necessary to
evaluate the number of neighbors to consider for a new input to the
model. This value for $k$ will determine if a company's stock is best described
by the known stock performance nearest to it, or by some double,
triple, or $k$-tuple averaging of nearest known stocks. The purpose,
then, is to derive an estimate for $k$ given a training set. This is
accomplished by performing a 10-fold cross-validation procedure on
partitions of the training set for values of $k \in \{1,\ldots,100\}$
and then choosing the $k^*$ for which the cross-validated error was first
minimized. For many industries, it was observed that $k^* = 1$ was the
most effective parameter for neighbor searches. Figure
\ref{fig:knn} provides a visualization of these measurements and
indicates the value of $k$ that first minimized the error. 

It was similarly crucial for the model to have some measure of when to
stop ``growing'' the random forest (by ``growing'' we suggest the
action of adding additional learned decision trees to the model). We
selected as the stopping criterion a point on the error curve for
which there is no substantial increase in accuracy by adding another
tree. Because of the variable and dynamic of these error curves, it is
necessary to evaluate such a criterion by inspection, so as to avoid
the pitfalls of local minima and premature termination of the
growing. 

Clearly, however, it is undesirable to have to refer to a ``true''
error rate, where the error is considered as a function of the number
of trees. Doing so would force the model to require a testing data set
for validation purposes, which is an unprincipled
approach. Fortunately, the random forest model offers a methodology
for obtaining an estimate of the error rate that is evaluated
internally at training. This error rate is referred to as the
Out-of-Bag (OOB) error and relies on the fact that the trees in the
forest are grown entirely on a bootstrapped sub-sample of the training
data. The essential idea is to evaluate all data points not used in
the training of an individual tree (about one-third of
$\vert\mathcal{D}_{\text{train}}\vert$) to obtain a classification
decision. For each data point, a naive voting system compares the
results of the forest to the true class and creates an average proportion of
the number of times the prediction is not consistent with the correct
class \cite{randomforest}. 

Figure \ref{fig:rf} presents an example of the OOB error estimate, the
true error curve created at testing, and a cross-validated error curve
using the entire data set. In this example, it was apparently the case
that the error estimate was an underestimate of the model's true
error. In practice, the OOB error tends to convincingly level out
around 120 grown trees, which consistently falls on a level portion of
the true error curve. This implies that this technique, while perhaps
growing a larger number of trees than necessary, will achieve a local
minimum of the true error as a function of the number of trees.

\subsection{Justifying the Boosting Procedure}

As mentioned in the relevant section on boosting the ensemble's
results in a committee fashion, the approach was unprincipled from a
theoretical perspective. That being said, it is possible to justify
the boosting from a empirical perspective. Below is a table
summarizing the individual results of each classification algorithm on
varying industry sectors in the period from the first quarter of 2009
to the second quarter of 2009. Training was performed on 10\% of the
available data for each industry. The selection of the 10\% of
stocks that constitute the training set were selected at random
from the total set of stocks. This implies that the size of the
training set varies by industry, since some industries are larger than others.


Furthermore, the training and testing error rates specified in the
table are reported \emph{post} boosting. The column \emph{Train}
corresponds to the error achieved by the model on the training data
set $\mathcal{D}_{\text{train}}$. Similarly, \emph{Test} is the error
rate reported by the model when deployed on the 90\% of the data not
used in training the model. The error rate presented here is
formulated as: 

\begin{align}
\text{Error Rate} = 1 - \frac{\text{\# of stocks correctly labeled}}{\text{\#
    of stocks considered}} 
\end{align}

\begin{table}[H]
\centering
\begin{tabular}{|l|l|l|l|l|l|l|}
\hline
\multicolumn{7}{c}{Error Rate for Individual Algorithms and Benchmarks}\\
\cline{1-7}
\emph{GICS Name} & \emph{Forest} & \emph{SVM} &
\emph{RVM} & \emph{$k$-NN Ensemble} & \emph{Train} & \emph{Test}\\\hline
Energy & 34.67\% & 43.34\% & 80.66\% & 58.39\% & 45.16\% & 19.34\%\\\hline
Materials & 34.01\% & 35.03\% & 71.43\% & 60.88\% & 33.33\% & 28.57\%\\\hline
Consumer Discretionary & 47.66\% & 47.66\% & 67.45\% & 49.14\% & 48.07\% & 32.55\%\\\hline
Consumer Staples & 48.10\% & 44.30\% & 65.19\% & 51.27\% & 38.89\% & 34.81\%\\\hline
Information Technology & 37.34\% & 37.18\% & 64.12\% & 49.03\% & 47.06\% & 35.88\%\\\hline
\end{tabular}
\caption{Reported here are the error rates reported for the constituent classifiers of
  the model, as well as the train set and test set errors for the
  ensemble classifier. Also reported is the test error of a naive
  committee model. Notice that the boosted model outperforms, or
  performs as well as, the algorithms considered individually. The
  performance of the individual algorithms on the training set is not
  reported chiefly because individual models often overfit to achieve
virtually zero training error, which is not meaningful.}
\label{tab:res}
\end{table}

The careful reader will have noticed that three GICS sectors were
excluded from Table \ref{tab:res}. In particular, the Industrials, Health Care, and
Financials industries were discovered to be difficult for the ensemble
to model effectively via a boosting operation described in Section
2.5. In these cases, a \emph{hypothesis} to explain the
uncharacteristic model behavior is that none of the algorithms individually perform
particularly weakly in these areas, typically scoring around a 30\%
error rate, which differs substantially from individual performances
observed in other industry sectors. Furthermore, we observed that in
the case of Health Care and Financials
shifting the decision threshold of the RVM from 0.8 to 0.5 resulted in
a dramatic increase in accuracy for these sectors, suggesting that a
high probability of a positive stock return is incompatible with these
stochastic industries. Results from the same time period as Table
\ref{tab:res} are reproduced below in Table \ref{tab:fail}:

\begin{table}[H]
\centering
\begin{tabular}{|l|l|l|l|l|l|l|}
\hline
\multicolumn{7}{c}{Individual Algorithms and Benchmarks (GICS 35 \& 40)} \\
\cline{1-7}
\emph{GICS Name} & \emph{Forest} & \emph{SVM} &
\emph{RVM} ({\tiny{threshold}}) & \emph{$k$-NN Ensemble} & \emph{Train} & \emph{Test}\\
\hline
Industrials & 33.05\% & 34.96\% & 31.78\% (0.8) & 33.05\% & 15.09\% & 31.78\%\\\hline
Health Care & 30.57\% & 36.09\% & 29.09\% (0.5) & 32.70\% & 7.69\% & 32.70\%\\\hline
Financials & 30.43\% & 36.23\% & 30.43\%  (0.5)& 27.53\% & 12.50\% & 30.43\%\\\hline
\end{tabular}
\caption{The error of the individual and ensemble algorithms for the
  Financials and Health Care industries. Notice that no individual
  algorithm performs weakly in these industries, which calls into
  question the necessity of boosting. In the case where no individual
  algorithm performed badly, it was characteristic of the boosting to
  learn to reproduce a single (not necessarily optimal) error rate of
  the constituent learners.}
\label{tab:fail}
\end{table}

\subsection{Results for Aggregated GICS Data}

A major component of this work was the division of market areas into
distinct sectors as specified by the GICS. This was motivated, in
part, by the relevance of particular explanatory variables to
particular fields of financial prediction. That being said, the model
is capable of being deployed in the context of a ``full market.'' In
other words, we can fail to specify divisions and accept all stocks as
not belonging to any particular field. 

This approach is not without its disadvantages, and the most stressing of these is algorithmic in nature. In particular, the
standard model, which was trained on the individual sectors, will on
the average, execute within 11.12 seconds (this includes all training,
cross-validation, and testing procedures). By contrast, the model
executed on the aggregated data is slower by nearly a factor of five,
completing its runtime in 53.46 seconds on the average. The
implementation of these algorithms was within the MATLAB environment,
suggesting that significant runtime improvements could be realized
with C or Fortran implementations; but despite these potential
advances, it is our suspicion that a significant gap in execution
time between the aggregated and partitioned models is unavoidable. 

It was found that the model trained on the stock data from the first
and second quarters of 2009
produced an accuracy of 39.34\%, which, while failing to outperform
the industry-specific classifications, would seem to indicate that an
aggregated approach to stock price return prediction is not
necessarily unreasonable. However, closer inspection of the individual
learning algorithms reveals that the RVM, and, to a certain extent, the
random forest, tremendously overfit the financial data and essentially
learn to reproduce the most common classification result. In the case
of the first and second quarters of 2009, this resulted in models that
had not actually learned from the data, producing instead a nearly
constant forecast of positive returns for all stock inputs. The high
accuracy then comes as a result of guessing the most common class
label. 

Indeed, users of the ensemble model should be wary of such a low
training error, which effectively means that the model has learned
\emph{nothing} except to memorize the input data. A potential
explanation for the model's decision to reproduce the results of the
$k$-Nearest Neighbors algorithm only stems from the fact that this
constituent model achieves essentially zero error on the training
set. In turn, the model will recognize the superiority of the
$k$-Nearest Neighbors approach and will adopt it without question. 

In light of these concerns over severe overfitting of the training
data, it is recommended not to apply this learning methodology to the
aggregated GICS sectors, but instead to partition them so as to avoid
the involved problems of learning essentially nothing from the data.

\begin{table}[H]
\centering
\begin{tabular}{|l|l|l|l|l|l|l|}
\hline
\multicolumn{7}{c}{Individual Algorithms and Benchmarks (2009 Q1 -
  2009 Q2)} \\
\cline{1-7}
\emph{GICS Name} & \emph{Forest} & \emph{SVM} &
\emph{RVM} & \emph{$k$-NN Ensemble} & \emph{Train} & \emph{Test}\\
\hline
Aggregated & 34.60\% & 49.61\% & 32.07\% & 39.20\% & 0.91\% & 39.20\%\\\hline
\end{tabular}
\caption{The error of the individual and ensemble algorithms for the
  aggregated industries. As is consistent with instances where no
  particular algorithm performs poorly, here it was observed that the
  boosting procedure essentially imitated the $k$-nearest neighbors
  solution. Interestingly, the RVM chose $\gamma = 1$ for its
  cross-validated hyperparameter, and the SVM chooses $\gamma =
  \frac{1}{2}$, suggesting that there is a subtle difference in
  optimal learning conditions by using the aggregated and partitioned
  financial data.}
\label{tab:agg}
\end{table}

\subsection{Results for Time-Series Financial Data}

It may be of some interest to attempt to forecast stock price returns in quarters far beyond the training instance.
 Here we
consider models trained from first quarter of 2006, and from the
first quarter of 2007, with the intent of accurately predicting the
immediate subsequent quarter. The parameters learned from this
operation are then used to predict the stock price return \emph{for
  all} subsequent quarters until the third quarter of 2012. Speaking
more formally, here we train four models corresponding to the random
forest, the SVM, the RVM, and a $k$-Nearest Neighbors ensemble. Each
model then learns its structure and parameters, and performs the
typical 5-fold cross validation procedure when necessary to confirm
the selection of hyperparameters. Then, these learned models are used
to form predictions for subsequent quarters following the training
procedure. However, these parameters are never relearned by the
models, but are kept constant in order to be consistent with the idea
that one cannot have significant prior knowledge of what will happen
definitely in the future. 

\begin{figure}[H]
        \centering
        \begin{subfigure}[b]{0.4\textwidth}
                \centering
                \includegraphics[width=\textwidth]{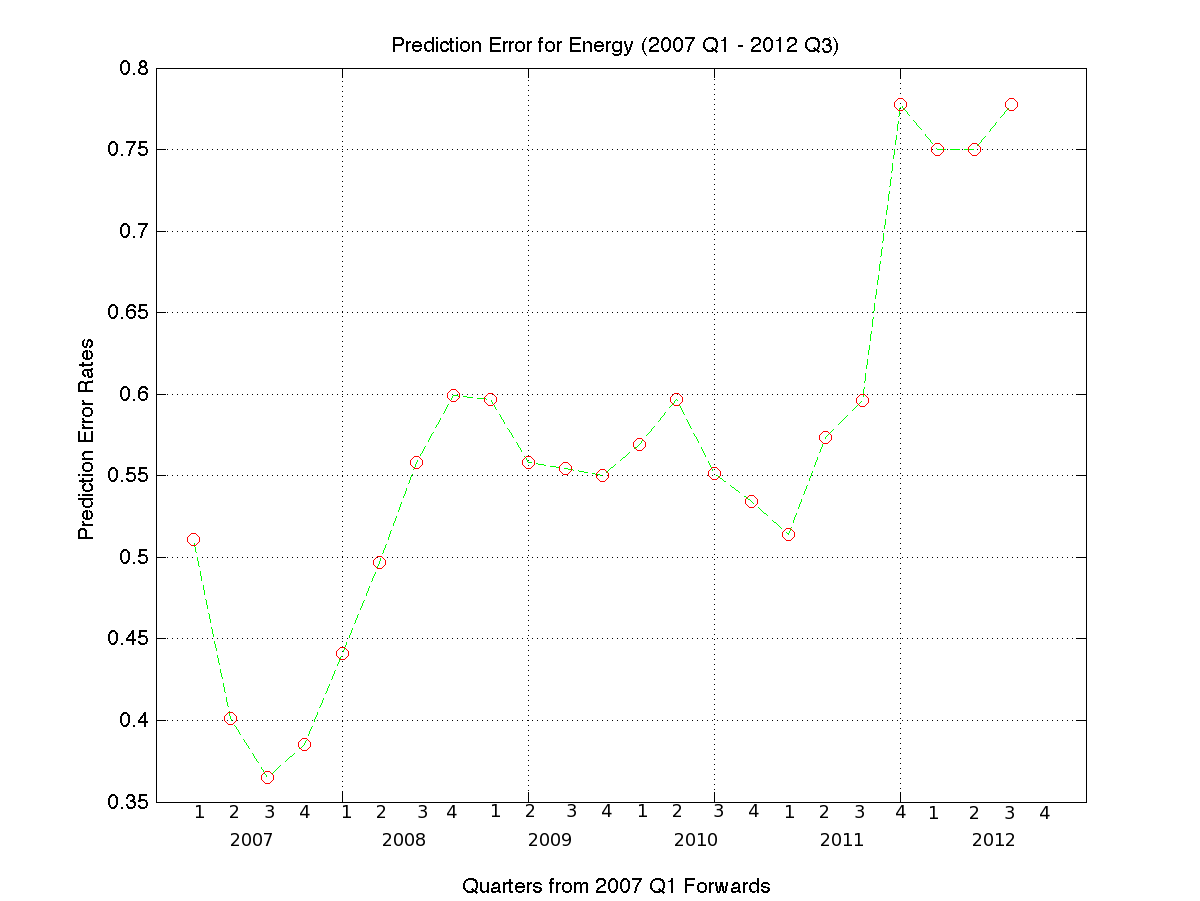}
                \caption{Energy finds a minimum error rate three
                  quarters after the initial training
                  procedure. Notice that the error rate fails to
                  surpass random guessing in some periods.}
                \label{fig:2007_10}
        \end{subfigure}%
        ~ 
        \begin{subfigure}[b]{0.4\textwidth}
                \centering
                \includegraphics[width=\textwidth]{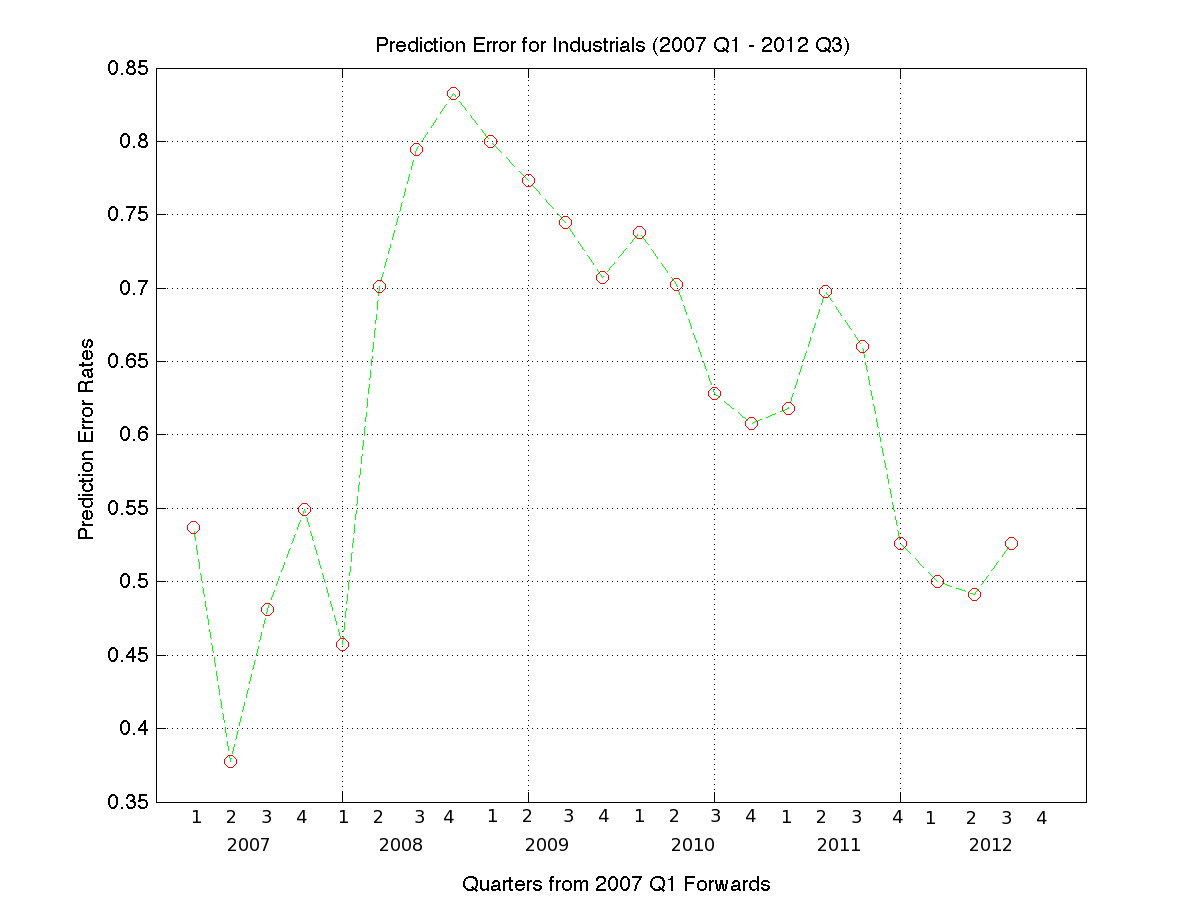}
                \caption{Industrials find a minimum error rate two
                  quarters after the initial training
                  procedure. Explanatory variables are expected to briefly
                  remain relevant.}
                \label{fig:2007_20}
        \end{subfigure}

        \begin{subfigure}[b]{0.4\textwidth}
                \centering
                \includegraphics[width=\textwidth]{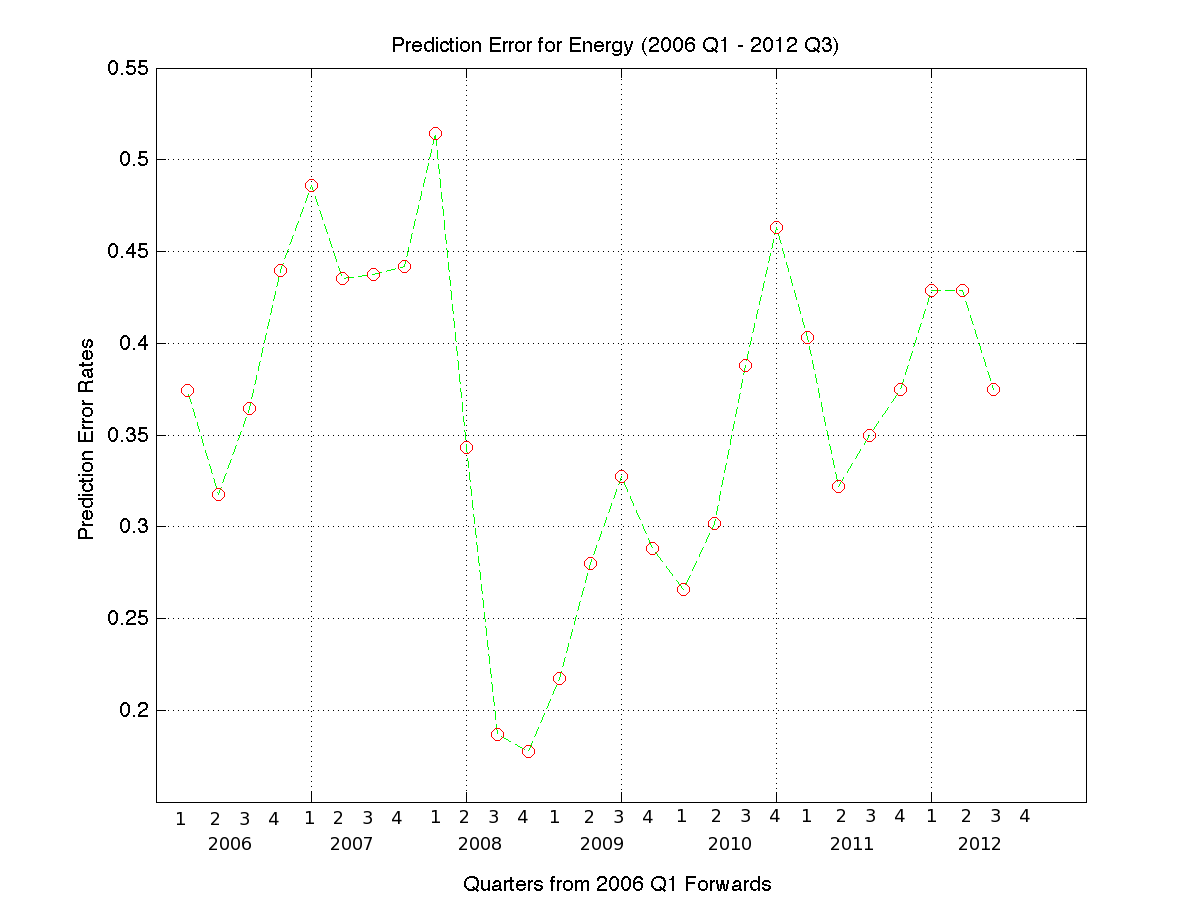}
                \caption{Energy demonstrating high
                  predictive accuracy eleven quarters after the training period.}
                \label{fig:2006_10}
        \end{subfigure}
        ~
        \begin{subfigure}[b]{0.4\textwidth}
                \centering
                \includegraphics[width=\textwidth]{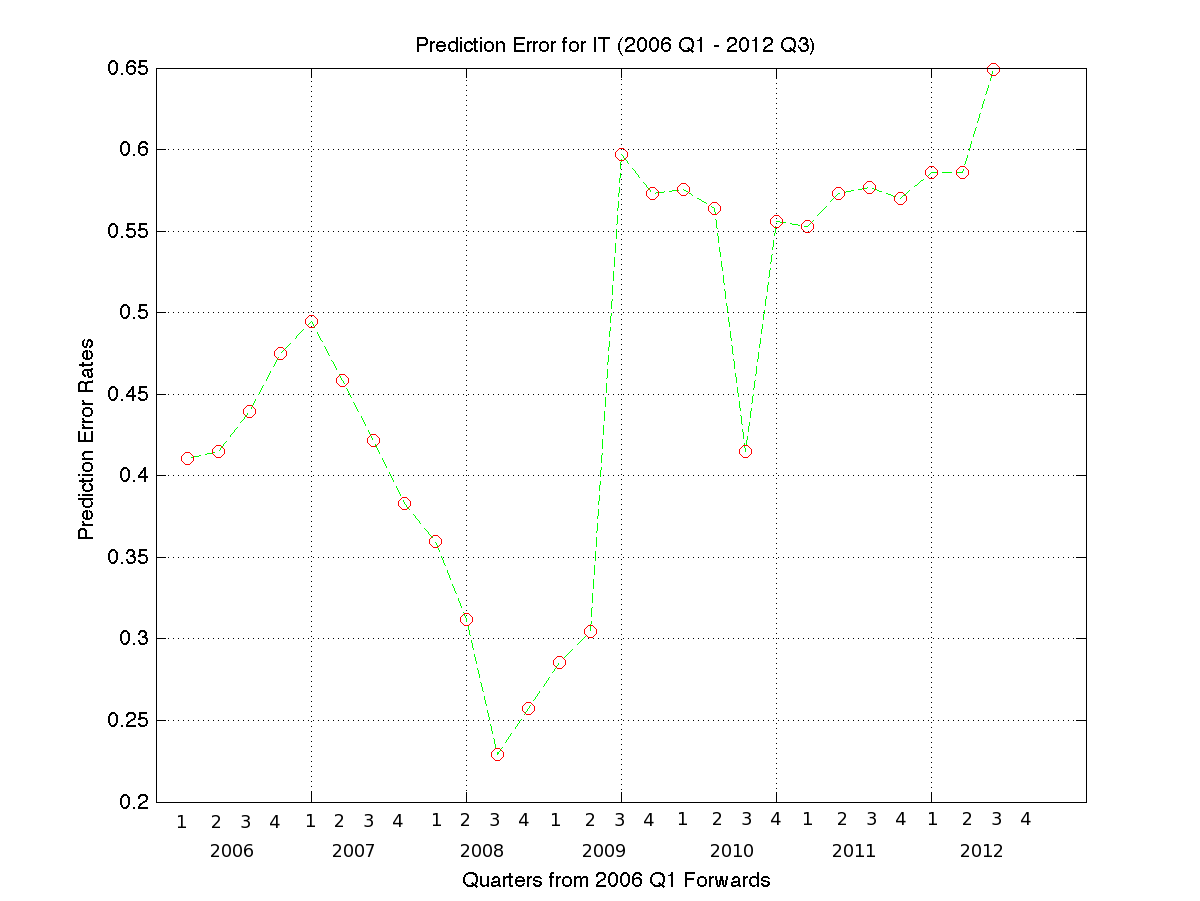}
                \caption{Information Technology demonstrating high
                  predictive accuracy twelve quarters after the training period.}
                \label{fig:2006_45}
        \end{subfigure}
        \caption{Presented here are stock return forecasts for both
          the first quarter of 2006 and 2007, and proceeding forwards
          to predict all subsequent quarters through the third
          quarter of 2012. In other words, models are learned for
          first two subsequent financial quarters, and these learned
          parameters are then \emph{applied} to data from the
          remaining time intervals. This is not equivalent to training
        an individual model to specifically predict the stock return
        from, for example, the first quarter of 2007 to the third
        quarter of 2008.}
        \label{fig:predictions}
\end{figure}

The empirical evidence we present involves training multiple,
industry-partitioned models on the years and quarters mentioned
previously. In the case of most industries, it is discovered that
training on a model on a given interval tended to produce good
accuracy for financial quarters immediately subsequent to the training
period. This behavior is characteristically observed, for example, in
Figures \ref{fig:2007_10} \& \ref{fig:2007_20}. Notice that in these
instances the error rate can sometimes exceed the random guessing
threshold in later quarters to achieve remarkably poor performance;
this is suggestive of a phenomenon where characteristics of the data
that are learned to be relevant in one quarter become distinctly
uninformative later on.


This phenomenon was not observed consistently throughout all
industries and time intervals, however. Certain industries and certain
time intervals sometimes displayed significant explanatory power over
stock price returns remarkably far in advance. This is demonstrated in
particular in the Energy and Information Technology when trained for
predicting the second quarter of 2007 with first quarter data (see
Figures \ref{fig:2006_10} \& \ref{fig:2006_45}). It is an interesting
observation to note that in both cases this remarkable increase in
predictive accuracy arrived in the 2008 year, when the financial
crisis caused significant distress in the stock market. The high
accuracy of the ensemble model in either case (around the 20\% error
rate mark) implies that a trader using this model to forecast stock
price returns far in advance would have been able to skillfully profit
from the market crash.

\begin{figure}[H]
        \centering

        \begin{subfigure}[b]{0.3\textwidth}
                \centering
                \includegraphics[width=\textwidth]{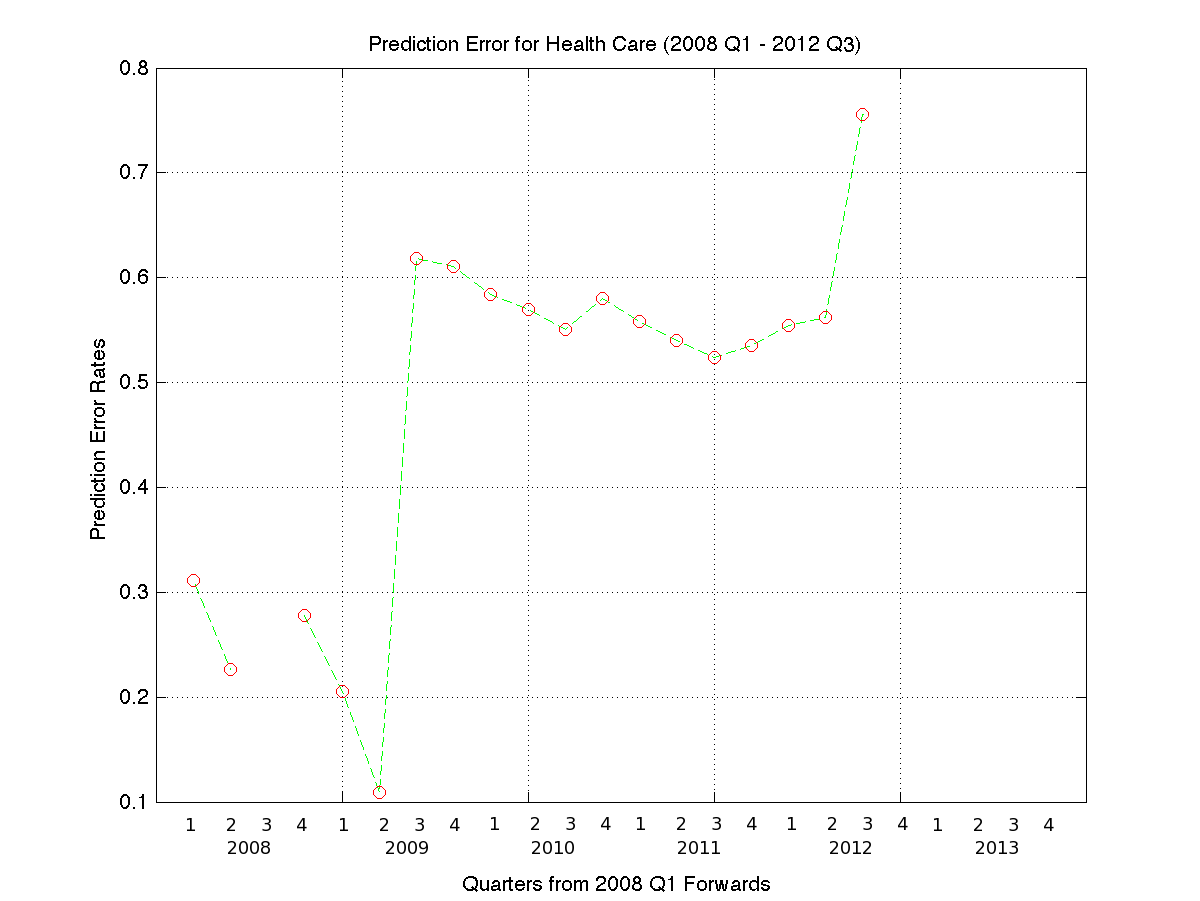}
                \caption{Consumer Discretionary}
                \label{fig:2008_25}
        \end{subfigure}
        ~
        \begin{subfigure}[b]{0.3\textwidth}
                \centering
                \includegraphics[width=\textwidth]{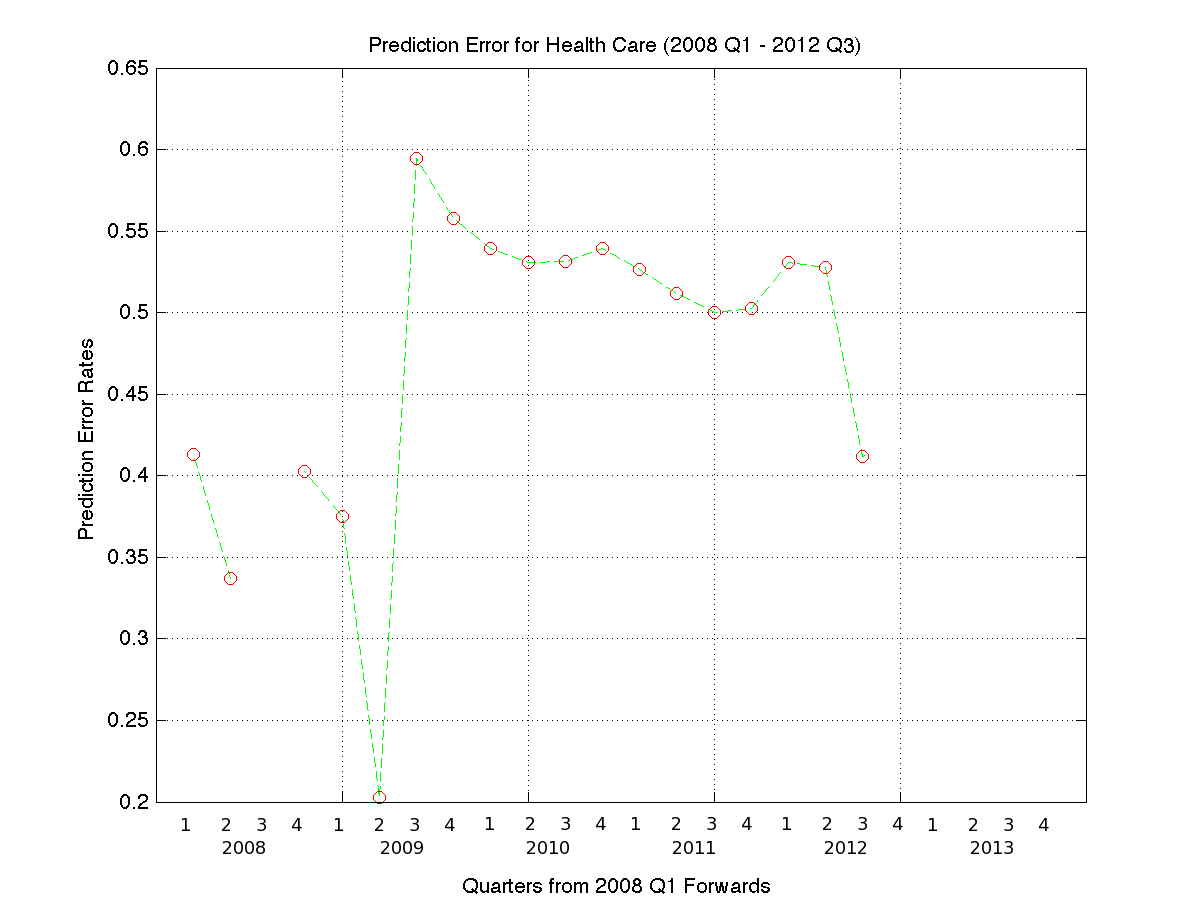}
                \caption{Health Care}
                \label{fig:2008_35}
        \end{subfigure}
        ~
        \begin{subfigure}[b]{0.3\textwidth}
                \centering
                \includegraphics[width=\textwidth]{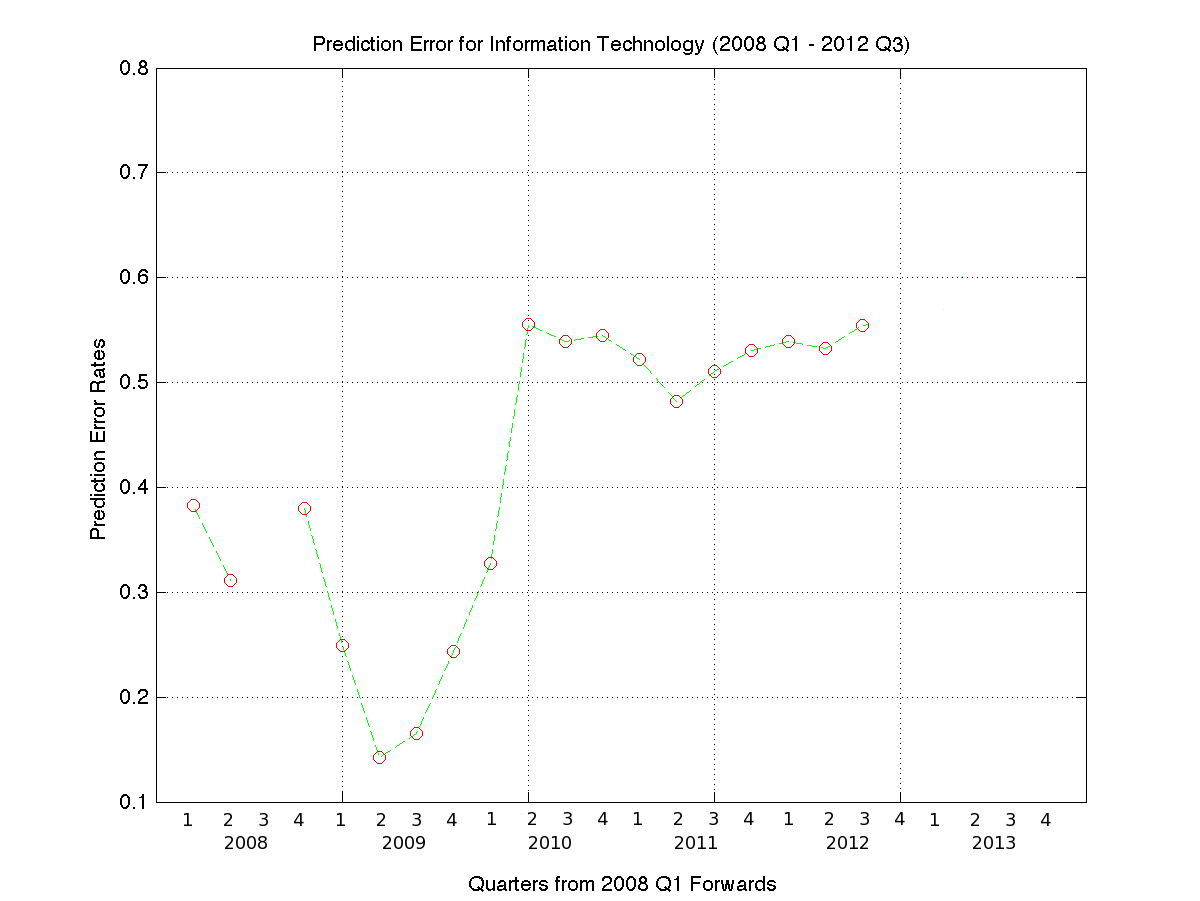}
                \caption{Information Technology}
                \label{fig:2008_45}
        \end{subfigure}
        \caption{We train the ensemble model on stock prices from the
          first two quarters of 2008 and proceed to predict stock
          returns for the remaining financial quarters until the third quarter
          of 2012. These visualizations suggest that the model does
          maintain explanatory power even in extremely uncertain and
          shaky financial circumstances, similar to those that
          surrounded the 2008 financial collapse.}
        \label{fig:predictions_2008}
\end{figure}

As a point of comparison, we also trained the model with the intention
of forecasting stock returns from 2008's first quarter to the second
quarter of the same year. These learned parameters were then used to
predict stock prices for all subsequent quarters in the same style as
the previous experiments. The key results of this analysis are
reproduced graphically in Figure \ref{fig:predictions_2008}. In
particular, we reproduce results from Consumer Discretionary, Health
Care, and Information Technology sectors. These visualizations suggest
that the ensemble algorithm is capable of predicting stock price
returns effectively when trained on data in the midst of financial
crises, maintaining an error rate of approximately 30\% in most
instances. In the interest of full disclosure, however, we note that
the algorithm \emph{failed} to learn anything effective for prediction
in the Financials sector, consistently failing to breach the random
guessing error benchmark of 50\%.

\section{Conclusions and Recommendations for Further Research}

We have presented an architecture for learning to predict stock price
returns by considering a binary classification problem, where positive
return predictions are denoted by the class label $+1$ and negative
predictions $-1$. This architecture relies on an ensemble committee
model of random forests, relevance vector machines, support vector
machines, and a $k$-nearest neighbor constituent ensemble. The
ensemble architecture we present has explanatory power over immediate
quarters that has been demonstrated to fall in the range of
approximately 70\% accuracy on testing data from financial quarters
between 2006 and 2012, inclusive. The model can
occasionally overfit the data, leading to poor performances on the
test set, but in practice this happens in a minority of possible
applications of the committee. 

Further research in the field of financial modeling should, of course,
by encouraged and pursued. To this end, we recommend
exploring the applications of so-called ``Deep Learning''
methodologies to stock price prediction. This chiefly involves
learning weight coefficients on a large directed and layered
graph. Models of this form have, in the past, proven difficult to
train and optimize, but recent breakthroughs in only the last several
months have lead to a renaissance in deep learning. 

More directly related to the research presented here is to incorporate a
more sophisticated weighting scheme in the boosting
procedure. Furthermore, it may be worthwhile to attempt representation
learning using autoencoders, rather than relying on pure input
data. This has the advantage of representing stocks not as collected
data as in Table \ref{tab:datatable}, but as a linear combination of ``basis stocks,'' which may be
discovered to capture interesting interrelationships that cannot be
perceived by normal methodologies. For further reading on the topic of
autoencoding algorithms, refer to \cite{auto}. Furthermore, it is of course preferable to be able to obtain stock
price return classifications on a daily basis. Accessing daily
time-series data on stock price returns encourages the prediction of
relatively returns immediately,
rather than models that predict months or years in
advance. Additionally, incorporating such financial phenomena as
earning surprises and additional explanatory variables would be
worthwhile for further refining the efficacy of the model.

\section{Acknowledgments}

The author would like to thank Professor Meifang Chu for her
valuable assistance in data collection and for general guidance in
advancing this project. This work was supported, in part, by a Neukom
 Institute grant for computational research.

\pagebreak[4]


\begin{thebibliography}{56}
\bibitem{bishop}
Bishop, Christopher M. \emph{Pattern Recognition and Machine
  Learning}. New York: Springer, 2006. Print.
\bibitem{randomforest}
Breiman, Leo, and Adele Cutler. ``Random Forests.'' UCB
Department of Statistics, Web. 22 Apr. 2013.
\bibitem{data}
Compustat Database. \emph{Wharton Research Data Services}. University of Pennsylvania, Web. 13 Apr. 2013. https://wrds-web.wharton.upenn.edu/wrds/.
\bibitem{huerta}
Huerta, Ramon, Elkan, Charles and Corbacho, Fernando, \emph{Nonlinear
  Support Vector Machines Can Systematically Identify Stocks with High
  and Low Future Returns} (September 6, 2012). 
\bibitem{auto}
H. Lee, A. Battle, R. Raina, and A. Y. Ng. \emph{Efficient sparse
  coding algorithms}. NIPS, 2007.
\bibitem{kim}
K. Kim, \emph{Financial Time Series Forecasting Using Support Vector
Machines}, Neurocomputing, 55, 307-319 (2003).
\bibitem{sewell}
M. V. Sewell, \emph{The Application of Intelligent Systems to Financial Time
Series Analysis}, Ph.D thesis, Department of Computer Science,
University College London, University of London (2012)
\bibitem{reliefadvantage}
Megchelenbrink, Wout, \emph{Relief-Based Feature Selection in
  Bioinformatics: Detecting Functional Specificity Residues from
  Multiple Sequence Alignments}, Master thesis, Department of
Information Science, Radboud University, Nijmegen (2010)
\bibitem{murphy}
Murphy, Kevin. \emph{Machine Learning A Probabilistic
  Perspective}. Cambridge: MIT, 2012. Print.
\bibitem{ranking}
N. Jegadeesh, and S. Titman, \emph{Returns to Buying Winners and
  Selling Losers: Implications for Stock Market Efficiency}, Journal
of Finance, 48 65-91, (1993).
\bibitem{tipping}
Tipping, Michael E. \emph{Sparse Bayesian Learning and the Relevance
  Vector Machine}. Journal of Machine Learning Research, 211-244
(2001).
\bibitem{relieff}
Wang, Yuhang, and F. Makedon. \emph{Application of Relief-F Feature
  Filtering Algorithm to Selecting Informative Genes for Cancer
  Classification using Microarray Data}.
\end{thebibliography}
\end{document}